\documentclass{article}

\usepackage[main,preprint]{neurips_2026}

\usepackage[utf8]{inputenc} %
\usepackage[T1]{fontenc}    %
\usepackage{hyperref}       %
\usepackage{url}            %
\usepackage{booktabs}       %
\usepackage{amsfonts}       %
\usepackage{nicefrac}       %
\usepackage{microtype}      %
\usepackage{xcolor}         %

\usepackage{graphicx}
\usepackage{subcaption}
\usepackage{amsmath}
\usepackage{amssymb}
\usepackage{mathtools}
\usepackage{xspace}
\usepackage{xparse}
\usepackage{pifont}
\usepackage{multirow}
\usepackage{enumitem}
\usepackage{listings}
\usepackage{wrapfig}
\usepackage{mdframed}
\usepackage{makecell}
\usepackage{float}
\usepackage{placeins}
\usepackage{colortbl}
\usepackage{capt-of}
\usepackage{cleveref}
\crefname{section}{Sec.}{Sec.}
\usepackage[most,skins,theorems]{tcolorbox}
\usepackage{algorithm}
\usepackage{algorithmic}
\usepackage{subcaption}
\usepackage{pgfplots}
\pgfplotsset{compat=1.18}
\usepackage{dsfont}
\providecommand{\mathbbm}[1]{\mathds{#1}}

\definecolor{darkblue}{rgb}{0, 0, 0.5}
\definecolor{darkgreen}{RGB}{50,100,0}
\definecolor{darkred}{RGB}{200, 0, 0}
\definecolor{lightblue}{RGB}{220,235,250}
\definecolor{lightgreen}{RGB}{220,250,220}
\definecolor{diffblue}{RGB}{0,0,0}
\definecolor{barGrey}{RGB}{169,169,169}
\definecolor{barRed}{RGB}{166,25,47}
\definecolor{flowTeal}{RGB}{28,103,134}
\definecolor{flowPaleGreen}{RGB}{238,248,234}
\hypersetup{colorlinks=true, citecolor=darkblue, linkcolor=darkblue, urlcolor=darkblue}
\DeclareRobustCommand{\changed}[1]{{\color{diffblue}#1}}

\usepackage[most]{tcolorbox}
\newtcolorbox{takeaway}{
  enhanced, breakable,
  colback=darkgreen!6, colframe=darkgreen,
  boxrule=0pt, leftrule=2.8pt,
  sharp corners=all,
  boxsep=0pt, left=8pt, right=8pt, top=4pt, bottom=4pt,
  before skip=5pt, after skip=6pt,
}

\newcommand{\lstbg}[3][0pt]{{\fboxsep#1\colorbox{#2}{\strut #3}}}

\lstdefinelanguage{diff}{
  basicstyle=\ttfamily\small,
  morecomment=[f][\lstbg{red!20}]-,
  morecomment=[f][\lstbg{green!20}]+,
}
\lstdefinelanguage{diffpython}{
  language=diff,
  morekeywords={def, if, else, for, while, return, import, from, as, class, with, try, except, finally, raise, lambda, and, or, not, in, is, None, True, False},
  morecomment=[l]{\#},
  morestring=[b]",
  morestring=[b]',
}

\tcbset{
  takeawaysbox/.style={
    title=Takeaways,
    colback=lightblue!80,
    colframe=black,
    fonttitle=\bfseries\small,
    coltitle=white,
    colbacktitle=black,
    enhanced,
    attach boxed title to top left={xshift=2.5mm,yshift=-2.5mm},
    boxed title style={rounded corners, size=small, colframe=black, colback=black},
    width=\linewidth,
    arc=3.5mm
  }
}

\NewDocumentCommand{\ganqu}{ mO{} }{\textcolor{blue}{\textsuperscript{\textit{ganqu}}\textsf{\textbf{\small[#1]}}}}
\NewDocumentCommand{\ybsun}{ mO{} }{\textcolor{magenta}{\textsuperscript{\textit{youbang}}\textsf{\textbf{\small[#1]}}}}

\newcommand{\method}{\textbf{T\textsuperscript{3}RL}\xspace}

\title{Tool Verification for Test-Time \\ Reinforcement Learning}

\author{%
  \begin{tabular}{c}
  Ruotong Liao$^{1,3,}$\thanks{Equal contribution.} \quad Nikolai R\"ohrich$^{1,}$\footnotemark[1] \quad Xiaohan Wang$^{2,}$\thanks{Correspondence: \texttt{liao@dbs.ifi.lmu.de}; \texttt{xhanwang@stanford.edu}.} \quad Yuhui Zhang$^2$\\[-1pt]
  Yasaman Samadzadeh$^1$ \quad Volker Tresp$^{1,3}$ \quad Serena Yeung-Levy$^{2,}$\footnotemark[2]\\[2pt]
  {\small\normalfont $^1$ Ludwig-Maximilians University of Munich \quad $^2$ Stanford University \quad $^3$ MCML}
  \end{tabular}
}

\begin{document}

\maketitle
\begingroup
  \renewcommand{\thefootnote}{}
  \footnotetext{Code: \url{https://github.com/mayhugotong/T3RL}}
\endgroup

\begin{abstract}

Test-time reinforcement learning (TTRL) has emerged as a promising paradigm for Recursive Self-Improving AI (RSI) by adapting Large Reasoning Models (LRMs) on unlabeled test inputs, using self-consensus rewards derived from majority voting over sampled rollouts.  However, majority voting can mistake popularity for correctness: a spurious yet high-frequency \emph{unverified consensus} may become a biased reward signal, causing test-time RL to reinforce frequent but wrong answers and collapse into an incorrect mode. We address this \emph{false-popular} failure mode with \method{} (\textbf{T}ool-Verification for \textbf{T}est-\textbf{T}ime \textbf{R}einforcement \textbf{L}earning), a verification-aware test-time RL framework. \method{} grounds pseudo-label construction in external tool evidence. Concretely, a verifier utilizes external tool evidence (e.g., from code execution) to upweight verified rollouts during a verification-aware voting, producing more reliable pseudo-labels for training. Across various math difficulties (MATH-500, AMC, and AIME 2024) and diverse backbone families, \method{} significantly improves over \textsc{TTRL}, with better performance on harder problems. \method{} is positioned as a \emph{verified online data synthesizer}, highlighting the role of tool verification in reliable online adaptation. \method{} is extensible to verifiable domains and provides training and inference efficiency.

\end{abstract}

\begin{figure}[H]
  \centering  \includegraphics[width=0.82\textwidth]{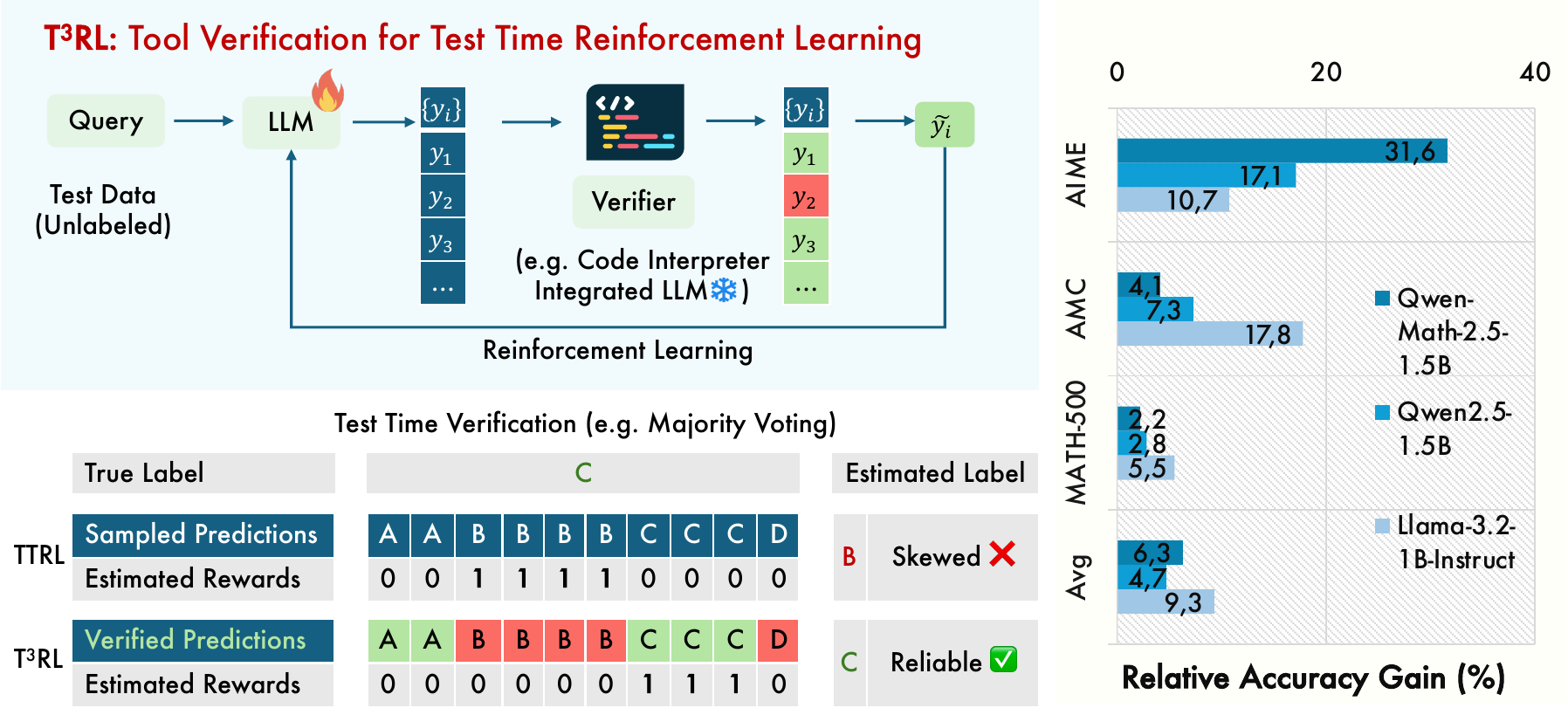}
  \caption{The concept of \method.
Top: \method{} introduces test-time verification into self-evolvement via tool-executed evidence (e.g., code interpreter) to stabilize training with verified rollouts. Bottom: Majority-vote pseudo-labels can be spurious. \method introduces verification to suppress false-popular pseudo-labels. Right: \method{} achieves consistent gains with evidence-grounded test-time training.}
  \label{fig:teaser}
\end{figure}

\section{Introduction}\label{sec:intro}

Modern language models often know more than they reliably say \citep{orgad2025llmsknowshowintrinsic, wang2023selfconsistencyimproveschainthought}. Pre-training and post-training endow models with broad factual and reasoning capabilities, yet a single inference-time generation may fail to express them. This gap has motivated \emph{test-time scaling} \citep{snell2024scalingllmtesttimecompute, muennighoff2025s1}: improving model performance by allocating additional computation at inference through sampling, search, verification, or aggregation \citep{snell2024scaling, wu2025inference, chang2025steplevelverifierguidedhybridtesttime, venktesh2025trust}. A key observation behind these methods is that, even when one rollout is unreliable, the distribution over many rollouts may contain a useful signal \citep{ICLR2026_776a5f2c, mekala2024toolverifier, lifshitz2025multiagentverificationscalingtesttime}.

\emph{Test-Time Reinforcement Learning} (TTRL) turns this signal into online learning \citep{zuo2025ttrl} for Recursive Self-Improving AI (RSI). Given an unlabeled test input, a model samples multiple reasoning traces, uses majority voting to infer a pseudo-label, and reinforces rollouts that agree with this majority answer. When the correct answer is the dominant mode of the model distribution, this provides a simple label-free reward signal that sharpens the model at test time. 

However, the majority answer is not always correct. If the model has a systematic bias, many rollouts may converge to the same wrong solution. In this case, voting selects popularity rather than truth: incorrect rollouts receive positive reward, correct minority rollouts are suppressed, and test-time RL amplifies the error. We call this failure mode \textbf{false-popular mode collapse}. As shown in \Cref{fig:teaser}, a frequent wrong answer can dominate the vote over a less frequent correct one, causing self-training to reinforce a false mode. Worse, once mode collapse occurs, it becomes even harder to self-correct.

This failure arises because self-consistency forms a closed loop: the model generates candidates, votes on them, and learns from its own vote. To break this loop, test-time learning needs evidence beyond the model's internal distribution. Many reasoning tasks provide exactly such evidence. Math solutions can be checked by computation, code can be executed against tests, and structured reasoning can be validated by external tools. These tasks are hard to solve but easy to verify.

We therefore ask: \textbf{Can label-free test-time reinforcement learning be made robust to false-popular mode collapse?} To answer this question, we propose \method{}, \textbf{Tool Verification for Test-Time Reinforcement Learning}. Rather than relying on majority vote alone, \method{} uses tools to verify candidate rollouts and constructs pseudo-labels through \emph{verification-weighted voting}. In short, the model should not only ask which answer is most frequent, but which answers can be verified.

Concretely, \method{} adds three components to TTRL. (1) A \emph{Verifier} extracts the final answer from each rollout and translates checkable reasoning steps into executable code. (2) A \emph{Verification Tool}, such as a Python interpreter, executes the code and returns evidence about the rollout's validity. (3) A \emph{Verification Weight} then gives verified rollouts greater vote mass when constructing the pseudo-label. The reward remains label-free, but is grounded in external tool feedback rather than raw self-consensus.

We evaluate \method{} on MATH500, AMC, and AIME2024. \method{} consistently improves over majority-vote TTRL, with larger gains on harder benchmarks and a maximum relative improvement of \textbf{31.6\%} on AIME2024 (\S\ref{sec:main_results}). We point out that \method{} is implicitly \textit{a verified synthetic data generator on the fly} (\S\ref{sec:data_syn}). Extensive ablations validate the importance of the \emph{Verifier}, the \emph{Verification Tool}, and the \emph{Verification Weight} (\S\ref{sec:abl}). We also demonstrate that \method{} is more robust (\Cref{fig:robustness}), test-time compute-efficient (\Cref{tab:compute}), and can be improved by stronger verifiers (\Cref{fig:ablation_two}).

More broadly, \method{} offers a plug-and-play framework for reliable test-time learning on verifiable tasks. While we instantiate it primarily with code-based verification for mathematical reasoning, our non-math Knights-and-Knaves experiment demonstrates that the same principle extends to ssymbolic verification with logical puzzles using SMT-LIB2 and Z3 (\S\ref{sec:domain_generalization}). We envision further extensions to code execution, theorem checkers, simulators, retrieval systems, and other task-specific verification tools.

\begin{figure*}[t]
  \centering
  \includegraphics[width=1\linewidth]{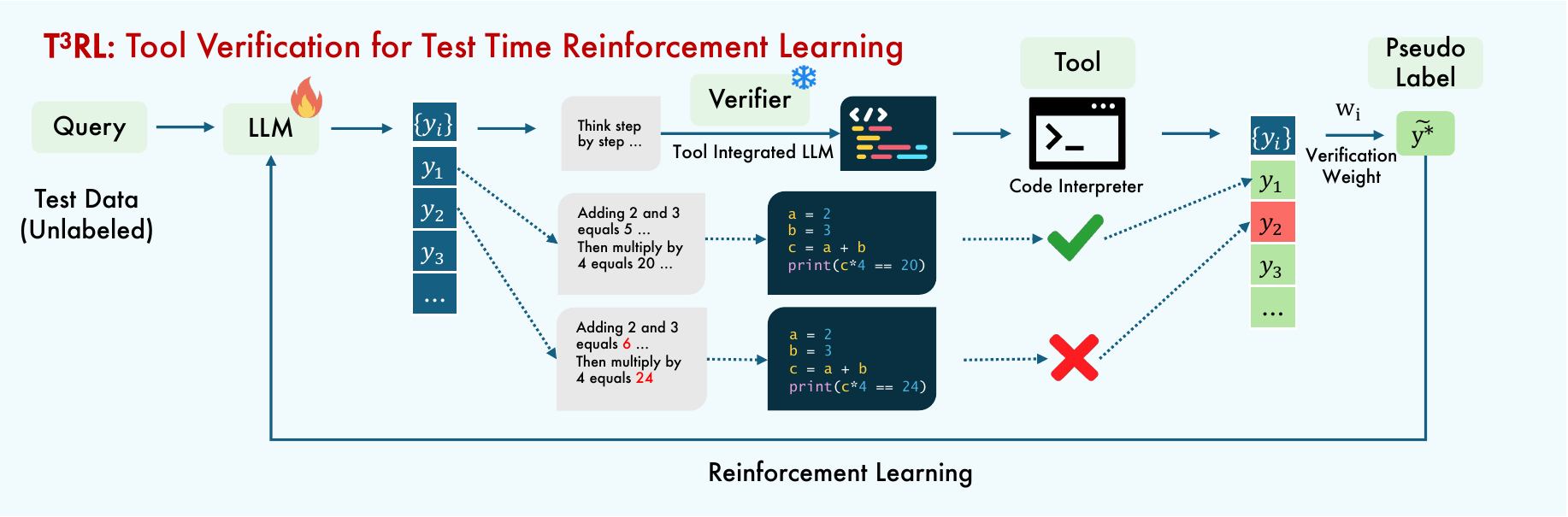}
\caption{\textbf{\method{}: Tool Verification for Test-Time Reinforcement Learning.}
\textit{Verifier:} an LLM verifier parses each sampled rollout $y_i$ into a candidate answer $a_i$ and, with tool feedback, produces a binary validity flag $v_i \in \{0,1\}$.
\textit{Tool verification:} the verifier compiles the rollout's claimed computations into lightweight Python and queries a code interpreter to obtain executable evidence.
\textit{Verification-weighted majority voting:} unverified rollouts receive unit vote mass, and verified rollouts receive $\omega$ vote mass. A verification-aware pseudo-label $\tilde{y}^{*}$ is formed by weighted voting.}
  \label{fig:main_fig}
\end{figure*}

\section{Related Works}

\paragraph{Verification for Test Time Scaling} 
Verification in test time scaling uses external verifiers to evaluate the quality of additional computation and select the best output from multiple candidates during inference. Verification mechanisms include reward models \citep{uesato2022solving, lightman2023let, cobbe2021training}, generative verifiers \citep{zhang2024generative}, symbolic checks \citep{ling2023deductive}, or multi-agent systems \citep{jin2025two, lifshitzmulti}. Recently, tool-integrated reasoning (TIR) \citep{gou2024toratoolintegratedreasoningagent} has brought a new perspective to tool-integrated verification \citep{mekala2024toolverifier, ICLR2026_776a5f2c}, formalizing tool use as an additional robust evidence. However, none of the prior works explore verification in \emph{test-time training}; we are the first to turn sampled rollouts into \emph{online, evidence-labeled} training instances through verification and shape the training process as a verified data synthesizer on-the-fly.

\paragraph{Test-Time Training}

Test Time Training (TTT) adapts model parameters during inference to handle distribution shifts on new tasks \citep{sun2020test, sun2024learning, Sun2024LearningT, behrouz2024titanslearningmemorizetest, liu2021ttt++}, in domains like video generation and understanding \citep{wang2025test, dalal2025one}, large language models \citep{hardt2023test}, or scientific discovery \citep{yuksekgonul2026learningdiscovertesttime}. Recently, TTT advanced to test-time reinforcement learning (TTRL) \citep{zuo2025ttrl} that combines unsupervised reinforcement learning  ~\citep{prasad2024self, zhang2025consistentpathsleadtruth} and reinforcement learning with verifiable rewards (RLVR) \citep{zeng2025simplerl, wang2025reinforcement}, which has been widely discussed in the era of self-evolving artificial intelligence. However, in the face of the challenge that spurious reward estimation presents to test-time training, none of the existing work has discussed verification. To the best of our knowledge, we are the first to propose test-time verification for test-time training, especially \textit{tool-assisted verification} for evidence-grounded test-time reinforcement learning. This provides a broader vision of agentic systems that increasingly rely on tool interaction, yet require reliable reward signals to for stable online learning.

\section{The Failure Mode: How Unverified Consensus Induces Reward Bias}

\subsection{Test Time Reinforcement Learning}
Unlike traditional RL, where models learn from known reward signals, TTRL operates on unlabeled test data without access to explicit supervision. TTRL is defined as follows:

Given a state represented by the prompt $x$, the model acts by producing an output $y$ sampled from a policy $\pi_\theta(y \mid x)$, parameterized by $\theta$. To construct a reward signal without ground-truth labels, TTRL generates multiple candidate outputs $\{y_1, y_2, \ldots, y_N\}$ through repeated sampling.
A consensus output $y^*$ is derived by \emph{majority voting}, serving as a proxy for the optimal estimation. The environment then provides a reward $r(y, y^*)$ based on the alignment between the sampled estimation $y$ and the consensus estimation $y^*$. The RL objective is thus to maximize the expected reward:
\begin{align}
\max_\theta \mathbb{E}_{y \sim \pi_\theta(\cdot \mid x)}[r(y, y^*)],
\label{eq:ttrl_objective}
\end{align}
and parameters $\theta$ are updated through gradient ascent:
\begin{align}
\theta \leftarrow \theta + \eta \nabla_\theta \mathbb{E}_{y \sim \pi_\theta(\cdot \mid x)}[r(y, y^*)],
\label{eq:ttrl_update}
\end{align}
where $\eta$ denotes the learning rate. 

\label{sec:ttrl_failure_mode}

\subsection{Spurious Majority as a Biased Pseudo-Label} 
\label{sec:false_popular}

\begin{figure}[t]
  \centering
  \begin{subfigure}[t]{0.48\linewidth}
    \centering
    \includegraphics[width=\linewidth]{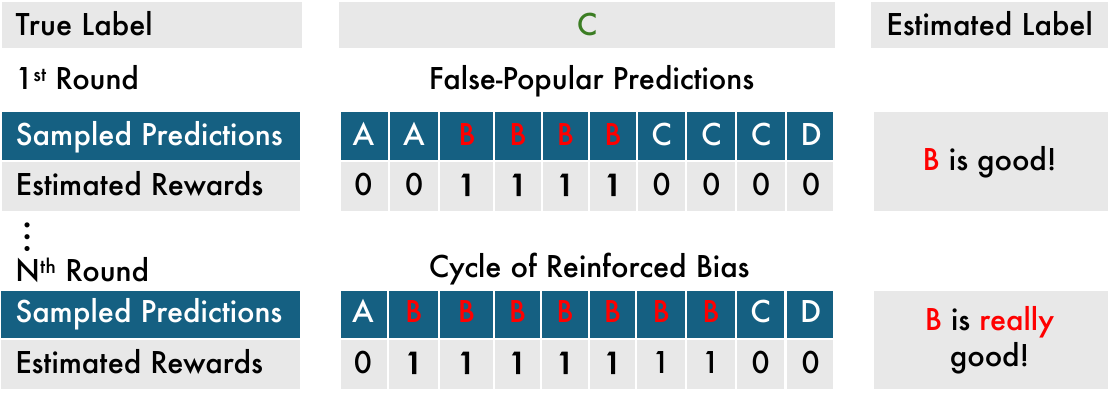}
    \caption{Failure case of \textsc{TTRL}.}
    \label{fig:false_popular}
  \end{subfigure}
  \hfill
  \begin{subfigure}[t]{0.48\linewidth}
    \centering
    \includegraphics[width=\linewidth]{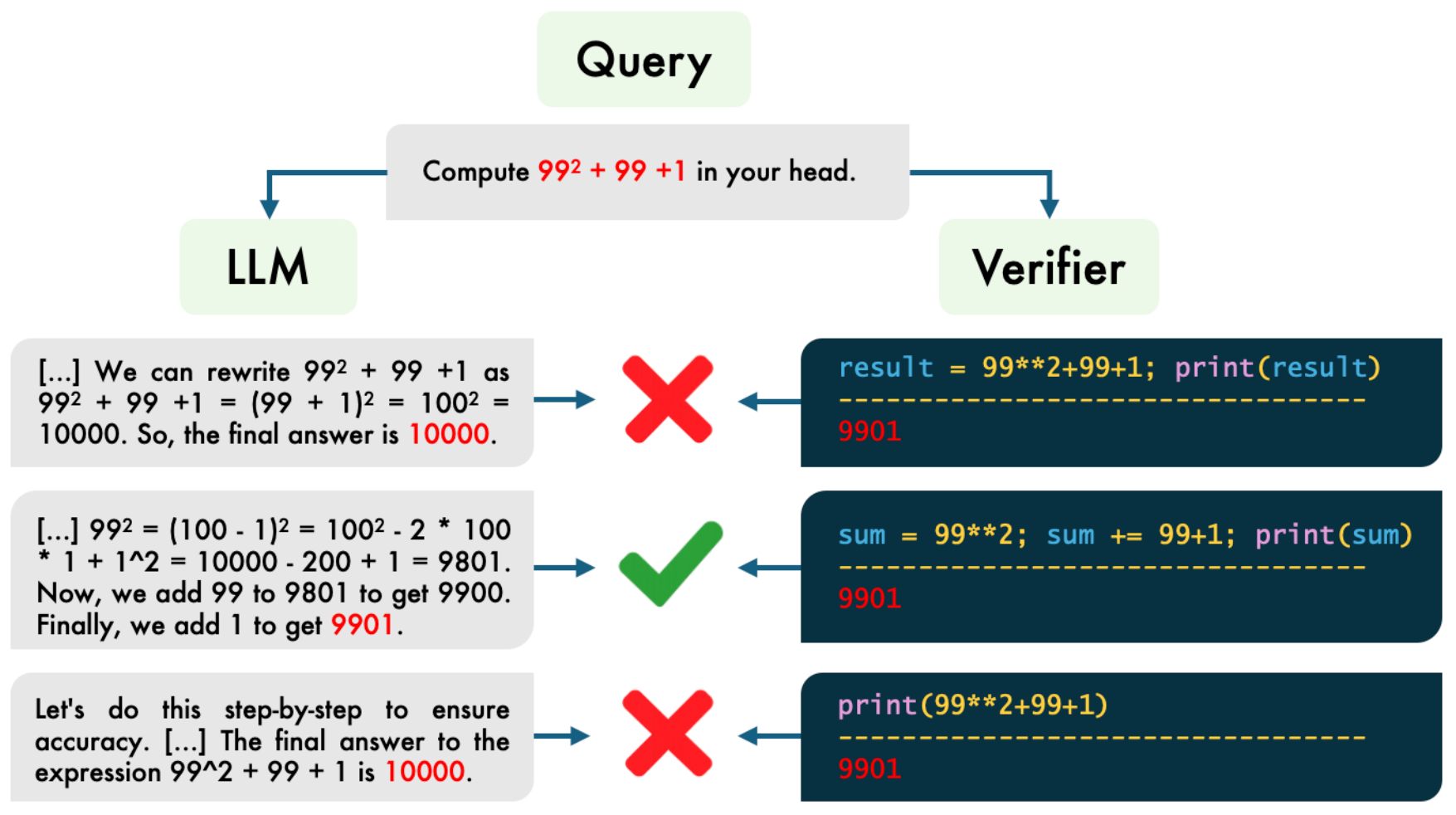}
    \vspace{-5mm}
    \caption{Success case of \method{}.}
    \label{fig:mvflip}
  \end{subfigure}
  \caption{\textbf{Unverified consensus versus tool-verified consensus.}
  (\subref{fig:false_popular})~Spurious reward in the reinforced cycle of \textsc{TTRL}: the false-popular answer is rewarded and its vote share grows over rounds.
  (\subref{fig:mvflip})~Tool verification can adjust the false-popular answer. }
  \label{fig:consensus_overview}
\end{figure}

\begin{wrapfigure}[10]{r}{0.36\linewidth}
\vspace{-\intextsep}
\centering
\begin{tikzpicture}
\begin{axis}[
  width=\linewidth, height=3.2cm,
  scale only axis=false,
  symbolic x coords={MATH-500,AMC,AIME 2024},
  xtick=data,
  ymin=15, ymax=85,
  ytick={20,40,60,80},
  yticklabel={\pgfmathprintnumber{\tick}\%},
  ylabel={False-majority share},
  ylabel style={font=\scriptsize, yshift=-2pt},
  tick label style={font=\scriptsize},
  every axis plot/.append style={very thick},
  grid=major,
  grid style={gray!25},
  axis line style={gray!60},
  enlarge x limits=0.18,
]
\addplot[
  color=diffblue,
  mark=*,
  mark options={fill=diffblue},
  point meta=y,
  nodes near coords={\pgfmathprintnumber[fixed,precision=2]{\pgfplotspointmeta}\%},
  every node near coord/.append style={font=\scriptsize, anchor=south, yshift=1pt},
]
  coordinates {(MATH-500,25.85) (AMC,46.07) (AIME 2024,73.33)};
\end{axis}
\end{tikzpicture}
\caption{\changed{\textbf{False-majority share} across benchmarks.}}
\label{fig:false_majority}
\vspace{-\intextsep}
\end{wrapfigure}

\paragraph{Self-consensus can estimate wrong labels.}
Let the generator induce a distribution over final answers, and consider two competing modes: the correct answer $y^\star$ and an incorrect but high-frequency answer $\tilde{y}$, e.g. \texttt{B} and \texttt{C} in \Cref{fig:false_popular} respectively.
If the win probability of \texttt{B} over \texttt{C} is non-zero, the majority vote can select $\tilde{y}$ as the pseudo-label. We measure the fraction of questions for which the unverified majority answer is wrong (\Cref{fig:false_majority}) and the fraction for which verification-weighted voting changes that answer (\Cref{fig:mvflip}). The false-majority share rises sharply with difficulty.

\paragraph{Self-reinforcing feedback loop and incorrect mode collapse.}
Once the pseudo-label is set to the false-popular $\tilde{y}$, the majority-based reward assigns positive reinforcement to rollouts that agree with the false signal $\tilde{y}$ and zero out rewards for truthful rollouts.
The RL update increases the likelihood of sampling $\tilde{y}$ in subsequent rollouts, which further increases its vote share, making the pseudo-label even more confidently wrong. This self-reinforcing dynamic can drive TTRL toward \emph{incorrect mode collapse}. Furthermore, once mode collapse occurs, the model becomes even harder to self-correct in an echo chamber.







\section{Method: Tool Verification for Test Time Reinforcement Learning}
\label{sec:method}

To prevent this failure mode, we present \method{}, an RL framework which integrates tool verification into the aggregation mechanism of test-time RL, thus achieving grounded and more robust reward estimation.
Specifically, \method{} has three core components:
(1) An external \textit{Verifier}, an LLM that is tasked with verifying a given reasoning trace by compiling the trace into executable Python code, and judging its validity based on the execution output (\S\ref{sec:method_verifier}); 
(2) The \textit{Verification Tool}, a code interpreter that executes the generated Python program and returns signals to the verifier (\S\ref{sec:method_tool});
(3) The \textit{Verification Weight}, which is used to replace the majority vote with a verification-aware weighted vote that boosts the voting power of verified rollouts. (\S\ref{sec:method_weight}).

\subsection{Verifier}
\label{sec:method_verifier}

A \emph{verifier} $\mathcal{V}$ is designed to evaluate each rollout before aggregation derives the estimated consensus label $y^*$:

\paragraph{Verifier.}
\changed{Given an input prompt $x$, we sample $N$ rollouts $\{y_i\}_{i=1}^{N}\sim \pi_\theta(\cdot\mid x)$ and extract the rollout's own final answer from each rollout,}
\begin{equation}
\hat{a}_i = \mathrm{Extract}(y_i), \qquad \hat{a}_i \in \mathcal{A}
\end{equation}
\changed{$\mathcal{V}$ then evaluates each rollout and returns only a validity flag}
\begin{equation}
(v_i) \;=\; \mathcal{V}(x, y_i,t_i),
\label{eq:verifier_triplet}
\end{equation}
\changed{where $v_i\in\{0,1\}$ indicates whether the rollout passes executable checks. $t_i$ represents the tool output from the verification tool $\mathcal{T}$ used by the verifier.

In \method{}, $\mathcal{V}$ is implemented as an LLM-based verifier that 
performs the following tasks: 1) generating the tool-calling query, by transforming the rollout rationale into a lightweight python program; 2) the execution of the \emph{verification tool} (\S\ref{sec:method_tool}), and 3) returning the verification validity results. Note that the verifier provides a reliability signal for $\hat{a}_i$.}

\subsection{Verification Tool}
\label{sec:method_tool}
A \emph{verification tool} $\mathcal{T}$ provides external, deterministic, and executable evidence to assist the verifier $\mathcal{V}$ by offloading specific verification tasks from $\mathcal{V}$ to an external tool.

\paragraph{Tool execution as external evidence.}
For math problems, many rollouts hinge on intermediate correctness (e.g., arithmetic computations, calculation hallucinations, etc).
In \method{}, \emph{tool verification} is initiated as a code interpreter $\mathcal{T}$, that is an executable checker that evaluates a verifier-generated Python program and returns the output of the executed code 
\begin{equation}
t_i \;=\; \mathcal{T}(\mathrm{Code}(x, y_i)).
\label{eq:tool_exec}
\end{equation}

\changed{The verifier then \emph{contrasts} the tool result with the rollout's extracted answer $\hat{a}_i$}
and produces a tool-verified validity indicator
\begin{equation}
v_i \;=\; \mathbbm{1}\!\left[t_i =\hat{a}_i\right].
\label{eq:tool_validity}
\end{equation}

\subsection{Verification Weight}
\label{sec:method_weight}

\paragraph{Heuristic Weighting.}
In standard TTRL, the consensus label is derived via simple majority voting, where every reasoning path contributes equally regardless of its logical soundness. However, in our framework, rollouts that pass the tool verification, i.e. for which $v_i=1$, are heuristically assumed to be more reliable than those that do not.
Since the ground truth is inaccessible at test time, we cannot calculate the exact probability of correctness for verified rollouts. Instead, we introduce a \emph{Verification Weight} hyperparameter, $\omega$, to quantify the voting power of tool-verified traces.

\paragraph{Verification-Aware Consensus.}
We modify the aggregation mechanism to use a verification-weighted majority vote. To maintain the diversity of rollouts, we ensure that each rollout maintains its contribution by setting the default weight to 1. For a set of rollouts, we assign a weight $w_i$ to each derived answer $\hat{a}_i$ based on the verification indicator $v_i$:
\begin{equation}
w_i \;=\; (1-v_i) \cdot 1 \;+\; v_i \cdot \omega,
\label{eq:weight_calc}
\end{equation}
where $\omega \geq 1$ is a fixed scalar hyperparameter. This ensures that unverified rollouts retain a unit vote, while each verified rollout contributes an upweight of $\omega$ votes.
The \emph{verification-aware} consensus label $\tilde{y}^{*}$ is then obtained by maximizing the total weighted vote mass:
\begin{equation}
\tilde{y}^{*}
\;=\;
\arg\max_{a\in\mathcal{A}}
\sum_{i=1}^{N} w_i \cdot \mathbbm{1}[\hat{a}_i=a].
\label{eq:verified_vote}
\end{equation}
This mechanism allows \method{} to shift the consensus derivation from the distribution of the most \emph{frequent} answer in the whole observed candidate set $\mathcal{A}$ towards the \emph{verified} consensus (see \Cref{fig:mvflip}), provided that the verified group possesses sufficient cumulative weight. 

\paragraph{Reward Calculation.}
Consistent with standard TTRL, the final reward signal remains binary but is now anchored to the robust, verification-aware consensus $\tilde{y}^{*}$. The reward for the $i$-th rollout is computed as:
\begin{equation}
r_i^{\text{v}}
\;=\;
\mathbbm{1}[\hat{a}_i=\tilde{y}^{*}].
\label{eq:t3rl_reward_binary}
\end{equation}

\paragraph{Training objective.}
We keep the TTRL update of Eq.~\eqref{eq:ttrl_update} unchanged in form, replacing the majority-vote pseudo-label $y^{*}$ with the verification-aware consensus $\tilde{y}^{*}$ of Eq.~\eqref{eq:verified_vote} and its induced reward, Eq.~\eqref{eq:t3rl_reward_binary}:
\begin{equation}
\max_{\theta}\;\mathbb{E}_{y\sim \pi_\theta(\cdot\mid x)}\!\left[r^{\text{v}}(y,\tilde{y}^{*})\right],
\qquad
\theta \leftarrow \theta + \eta \nabla_\theta \mathbb{E}_{y\sim \pi_\theta(\cdot\mid x)}\!\left[r^{\text{v}}(y,\tilde{y}^{*})\right],
\label{eq:t3rl_objective}
\end{equation}
where $\tilde{y}^{*}$ is held fixed within each update, and we optimize this objective with GRPO in practice.

\section{Experiments}\label{sec:experiments}

\subsection{Experimental Setup}

\paragraph{Benchmarks}
We evaluate \method on
$3$ mathematical reasoning benchmarks: AIME 2024~\citep{li2024numinamath}, AMC~\citep{li2024numinamath}, and MATH-500~\citep{hendrycks2021measuring}.

\paragraph{Models.}

We evaluate \method{} under the \emph{test-time reinforcement learning}
setting across diverse backbone configurations spanning model families,
alignment regimes, and scales (1B--4B), which appeals to the call to
validate methods on diverse models~\citep{shao2025spuriousrewardsrethinkingtraining}.
Our experiments cover both \emph{base} and \emph{instruction-tuned}
models, as well as \emph{math-specialized} backbones. Concretely, we study
LRMs from the Qwen and LLaMA families:
(1) \textbf{Vanilla:} \texttt{Qwen2.5-1.5B}~\citep{qwen2.5} and
\texttt{Qwen-3-4B}~\citep{qwen3};
(2) \textbf{Math:} \texttt{Qwen2.5-Math-1.5B}~\citep{qwen2.5};
(3) \textbf{Instruct:} \texttt{Llama-3.2-1B-Instruct} and
\texttt{Llama-3.2-3B-Instruct}~\citep{grattafiori2024llama3herdmodels}.

\begin{table*}[t]
  \caption{\textbf{Main performance of \method.} \label{tab:main_results}
  (a) Improvement across benchmark and backbone models. (b) Difficulty breakdown performance on MATH-500 (L5 is the hardest). The relative improvement of \method{} over TTRL increases with difficulty level (\Cref{fig:trend_sub_c}).}
  \centering

  \begin{minipage}[t]{0.45\textwidth}\vspace{0pt}
    \centering
    \small
    \renewcommand{\arraystretch}{1.05}

    {\captionsetup{type=table}\caption*{\changed{(a) Main results}}}
        
    \resizebox{\linewidth}{!}{
      \begin{tabular}{l|cccc}
      \toprule
      \textbf{Model / Method} & \textbf{AIME 2024} & \textbf{AMC} & \textbf{MATH-500} & \textbf{Avg} \\
      \midrule
      \multicolumn{5}{c}{\textbf{Math Model}} \\
      \midrule
      Qwen-2.5-Math-1.5B & 7.7 $\pm$ 2.0 & 28.6 $\pm$ 4.5 & 32.7 $\pm$ 2.7 & 23.0 $\pm$ 1.9 \\
      \quad w/ TTRL & 15.8 $\pm$ 2.1 & 48.9 $\pm$ 1.6 & 73.0 $\pm$ 3.8 & 45.9 $\pm$ 1.5 \\
      \rowcolor{lightgreen!25}\quad \textbf{w/ \method{}} & \textbf{20.8 $\pm$ 1.6} & \textbf{50.9 $\pm$ 2.7} & \textbf{74.6 $\pm$ 3.2} & \textbf{48.8 $\pm$ 1.5} \\
      \midrule
      \multicolumn{5}{c}{\textbf{Vanilla Model}} \\
      \midrule
      Qwen-2.5-1.5B & 0.2 $\pm$ 1.2 & 0.6 $\pm$ 2.2 & 7.7 $\pm$ 2.0 & 2.8 $\pm$ 1.1 \\
      \quad w/ TTRL & 3.5 $\pm$ 1.2 & 28.6 $\pm$ 4.5 & 63.2 $\pm$ 3.5 & 31.8 $\pm$ 1.9 \\
      \rowcolor{lightgreen!25}\quad \textbf{w/ \method{}} & \textbf{4.1 $\pm$ 1.2} & \textbf{30.7 $\pm$ 3.7} & \textbf{65.0 $\pm$ 2.1} & \textbf{33.3 $\pm$ 1.5} \\
      \midrule
      Qwen-3-4B & 0.0 $\pm$ 0.0 & 12.5 $\pm$ 4.4 & 50.4 $\pm$ 3.5 & 21.0 $\pm$ 1.9 \\
      \quad w/ TTRL & 36.4 $\pm$ 1.9 & 71.7 $\pm$ 3.0 & 88.4 $\pm$ 1.4 & 65.5 $\pm$ 1.3 \\
      \rowcolor{lightgreen!25}\quad \textbf{w/ \method{}} & \textbf{40.0 $\pm$ 1.6} & \textbf{74.2 $\pm$ 2.8} & \textbf{89.5 $\pm$ 1.4} & \textbf{67.9 $\pm$ 1.2} \\
      \midrule
      \multicolumn{5}{c}{\textbf{Instruct Model}} \\
      \midrule
      Llama-3.2-1B-Instruct & 0.8 $\pm$ 0.0 & 4.2 $\pm$ 2.6 & 4.4 $\pm$ 1.5 & 3.1 $\pm$ 1.0 \\
      \quad w/ TTRL & 7.5 $\pm$ 2.5 & 16.9 $\pm$ 1.6 & 40.0 $\pm$ 1.6 & 21.5 $\pm$ 1.1 \\
      \rowcolor{lightgreen!25}\quad \textbf{w/ \method{}} & \textbf{8.3 $\pm$ 2.4} & \textbf{19.9 $\pm$ 0.9} & \textbf{42.2 $\pm$ 2.6} & \textbf{23.5 $\pm$ 1.2} \\
      \midrule
      \changed{Llama-3.2-3B-Instruct} & 6.0 $\pm$ 1.4 & 19.4 $\pm$ 1.8 & 43.9 $\pm$ 2.1 & 23.1 $\pm$ 1.0 \\
      \quad w/ TTRL & 13.3 $\pm$ 3.8 & 31.3 $\pm$ 4.0 & 61.6 $\pm$ 5.1 & 35.4 $\pm$ 2.5 \\
      \rowcolor{lightgreen!25}\quad \textbf{w/ \method{}} & \textbf{17.1 $\pm$ 1.2} & \textbf{34.2 $\pm$ 2.7} & \textbf{63.3 $\pm$ 3.3} & \textbf{38.2 $\pm$ 1.5} \\
      \bottomrule
      \end{tabular}
    }

  \end{minipage}
  \hfill
  \begin{minipage}[t]{0.53\textwidth}\vspace{0pt}
    \centering
    \small
    \renewcommand{\arraystretch}{1.05}

    {\captionsetup{type=table}\caption*{(b) MATH-500 difficulty breakdown.}}

    \resizebox{\linewidth}{!}{
      \begin{tabular}{l|ccccc}
      \toprule
       & L1 & L2 & L3 & L4 & L5 \\
      \midrule
      Qwen-2.5-Math-1.5B & 25.9 & 33.0 & 36.3 & 32.5 & 22.3 \\
      w/ TTRL     & 95.0 & 88.0 & 85.0 & 69.7 & 47.0 \\
      \rowcolor{lightgreen!25}\textbf{w/ \method{}}
               & \textbf{95.2} & \textbf{88.2} & \textbf{87.3} & \textbf{72.0} & \textbf{49.0} \\
      \midrule
      Rel.\ (\% over Baseline)
               & $\uparrow$267.6\% & $\uparrow$167.3\% & $\uparrow$140.5\% & $\uparrow$121.5\% & $\uparrow$119.7\% \\
      Rel.\ (\% over TTRL)
               & \textbf{$\boldsymbol{\uparrow}$0.2\%} & \textbf{$\boldsymbol{\uparrow}$0.2\%} & \textbf{$\boldsymbol{\uparrow}$2.7\%} & \textbf{$\boldsymbol{\uparrow}$3.3\%} & \textbf{$\boldsymbol{\uparrow}$4.3\%} \\
      \bottomrule
      \end{tabular}
    }

    \vspace{0pt}

    \captionsetup{type=figure}
    \captionof{figure}{Relative gains across MATH-500 levels.}
    \label{fig:trend_sub_c}

    \begin{tikzpicture}
      \pgfplotsset{
        mathlevels/.style={
          width=159pt, height=55pt, scale only axis=true,
          symbolic x coords={MATH-L1,MATH-L2,MATH-L3,MATH-L4,MATH-L5},
          enlarge x limits=0.12,
          tick label style={font=\tiny},
          label style={font=\tiny},
        },
      }
      \begin{axis}[
        name=lvax, mathlevels,
        ybar stacked, bar width=11pt,
        xtick=data, xtick align=outside,
        ymin=105, ymax=307,
        ytick={150,200,250,300},
        ylabel={Improvement (\%)}, ylabel near ticks,
        ylabel style={font=\tiny, yshift=-1pt},
        ymajorgrids, grid style={gray!20, very thin},
        legend style={
          at={(0.98,0.96)}, anchor=north east, font=\tiny,
          draw=black, line width=0.3pt, fill=white,
          inner xsep=2pt, inner ysep=1pt, legend columns=2,
          /tikz/every even column/.append style={column sep=2pt},
        },
        legend image code/.code={\draw[#1] (0pt,-1.4pt) rectangle (6pt,1.4pt);},
      ]
        \addplot[fill=barGrey, draw=barGrey!70!black, line width=0.2pt]
          coordinates {(MATH-L1,266.8) (MATH-L2,166.7) (MATH-L3,134.2) (MATH-L4,114.5) (MATH-L5,110.8)};
        \addlegendentry{TTRL}
        \addplot[fill=barRed, draw=barRed, line width=0.2pt]
          coordinates {(MATH-L1,0.8) (MATH-L2,0.6) (MATH-L3,6.3) (MATH-L4,7.1) (MATH-L5,9.0)};
        \addlegendentry{\method{}}
      \end{axis}
      \begin{axis}[
        mathlevels,
        at={(lvax.south west)}, anchor=south west,
        axis x line=none, axis y line*=right,
        ymin=0, ymax=16, ytick={0,4,8,12,16},
        ylabel={Additional Gain (\%)}, ylabel near ticks,
        ylabel style={font=\tiny, yshift=1pt},
      ]
        \addplot[barRed, line width=0.6pt, mark=*, mark size=0.8pt,
                 mark options={fill=barRed, draw=barRed}]
          coordinates {(MATH-L1,0.8) (MATH-L2,0.6) (MATH-L3,6.3) (MATH-L4,7.1) (MATH-L5,9.0)};
        \node[font=\tiny\bfseries, above=0.5pt] at (axis cs:MATH-L1,0.8) {+0.8\%};
        \node[font=\tiny\bfseries, above=0.5pt] at (axis cs:MATH-L2,0.6) {+0.6\%};
        \node[font=\tiny\bfseries, above=0.5pt] at (axis cs:MATH-L3,6.3) {+6.3\%};
        \node[font=\tiny\bfseries, above=0.5pt] at (axis cs:MATH-L4,7.1) {+7.1\%};
        \node[font=\tiny\bfseries, above=0.5pt] at (axis cs:MATH-L5,9.0) {+9.0\%};
      \end{axis}
    \end{tikzpicture}
  \end{minipage}

    \vspace{2pt}
\end{table*}
\paragraph{Evaluation Setup}

We apply \method to each benchmark utilizing a maximum token limit of $2{,}560$. For our \textit{main experiments}, we follow the DeepSeek-R1~\citep{guo2025deepseek} protocol: we evaluate each experiment $4$ times (temperature $0.6$, top-$p$ $0.95$) and report pass@1 as the average correctness, formalized as:
\changed{For AIME 2024, whose test set is small, we average $16$ independent evaluations; for AMC and MATH-500 we average $4$. The reported $\pm$ values are standard deviations aross runs.}
\[
\text{pass@1} = \frac{1}{k} \sum_{i=1}^{k} p_i,
\]
where $p_i$ indicates if response $i$ is correct.

\paragraph{Baselines}

We compare our method against base models, as well as against TTRL \citep{zuo2025ttrl}, where we report the original results where possible. For experiments where TTRL results are not available, we reproduce training runs using the implementation details in \citet{zuo2025ttrl}.

\paragraph{Implementation Details}
We implement \method{} using GRPO \citep{shao2024deepseekmath} on each benchmark, training the policy model via AdamW with a cosine schedule (peak learning rate: $5 \times 10^{-7}$). To balance performance with computational efficiency, we follow TTRL \citep{zuo2025ttrl} in employing a vote-then-sample strategy: 64 responses are initially generated for label estimation (temperature 0.6) and downsampled to 32 for training. The maximum token length is 2,560. Regarding the verifier model, we find that we can restrict the generation length to $1{,}024$ tokens, since code snippets are relatively short. We also use a temperature of $0.6$ for the verifier across all benchmarks. Training runs are conducted on 8 NVIDIA A100 GPUs, with varying episode counts based on dataset size: 10 (MATH-500), 30 (AMC), and 80 (AIME 2024). The pseudo code is listed in Appendix~\ref{app:code} and the system prompt is in Appendix~\ref{appx:verifier_prompt}.

\subsection{Main Results}\label{sec:main_results}

\Cref{tab:main_results} shows that \method{} repeatedly outperforms \textsc{TTRL} across all evaluated models and benchmarks, supporting our central hypothesis that verification-shaped majority voting mitigates unverified-consensus bias during test-time RL.
\changed{Across the $15$ paired \method{}--TTRL model--benchmark comparisons, \method{} wins all $15$.}

\paragraph{Consistent gains across benchmarks and model types.}
\method{} improves over \textsc{TTRL} on all three evaluated benchmarks, spanning an easy-to-hard spectrum from MATH-500 (easiest), to AMC (medium), to AIME 2024 (hardest).
\method{} consistently improves over \textsc{TTRL} across five backbones
spanning three regimes (math, vanilla, instruction-tuned) and two scales
(1B--4B): the largest gains are on AIME 2024 for \texttt{Qwen-Math-1.5B}
($+31.6\%$) and \texttt{Qwen2.5-1.5B} ($+17.1\%$), and on AMC for
\texttt{Llama-3.2-1B-Instruct} ($+17.8\%$). The benefit is tied with three standing out trends:

\paragraph{Math-specialised backbones, harder benchmarks, and larger models
benefit most.}
\begin{itemize}
    \item \textit{(i) Math-specialised models benefit more.}
    \texttt{Qwen-Math-1.5B} achieves a larger average gain over
    \textsc{TTRL} than the vanilla \texttt{Qwen2.5-1.5B}
    ($+6.3\%$ vs.\ $+4.7\%$): math backbones generate longer calculation
    chains whose small arithmetic slips are precisely the kind of error
    that executable checks can catch.
    \item \textit{(ii) Harder benchmarks benefit more.} Within each
    backbone, gains peak on AIME 2024 (e.g., $+31.6\%$ for
    \texttt{Qwen-Math-1.5B}) and on the hardest MATH-500 level L5: longer
    computation chains accumulate errors that deterministic tool execution
    can still verify, preserving signal where self-consensus degrades.
    \item \textit{(iii) Gains scale to larger backbones.} On
    \texttt{Llama-3.2-3B-Instruct} AIME rises from $13.3$ to $17.1$
    ($+28.6\%$); on the much stronger \texttt{Qwen-3-4B} where \textsc{TTRL} already achieves $36.4$, it reaches $40.0$ ($+9.9\%$). The gain is largely preserved under self-verification ($+9.3\%$ / $+27.1\%$,
    \Cref{tab:self_verifier}), so the contribution scales with backbone
    capability rather than coming from external-coder distillation.
\end{itemize}

\subsection{Ablation Studies} \label{sec:abl}

This section presents an analysis of the three key factors that support \method: (1) test-time verification, (2) tool-assisted verification, and (3) verification-weighted majority voting. Unless otherwise specified, we run ablations on \texttt{Qwen-Math-1.5B} with rollout size $N{=}64$ and report Pass@1 performance.

\begin{figure}[h]
  \centering
  \begin{subfigure}[t]{0.495\columnwidth}
    \centering
    \includegraphics[width=\linewidth]{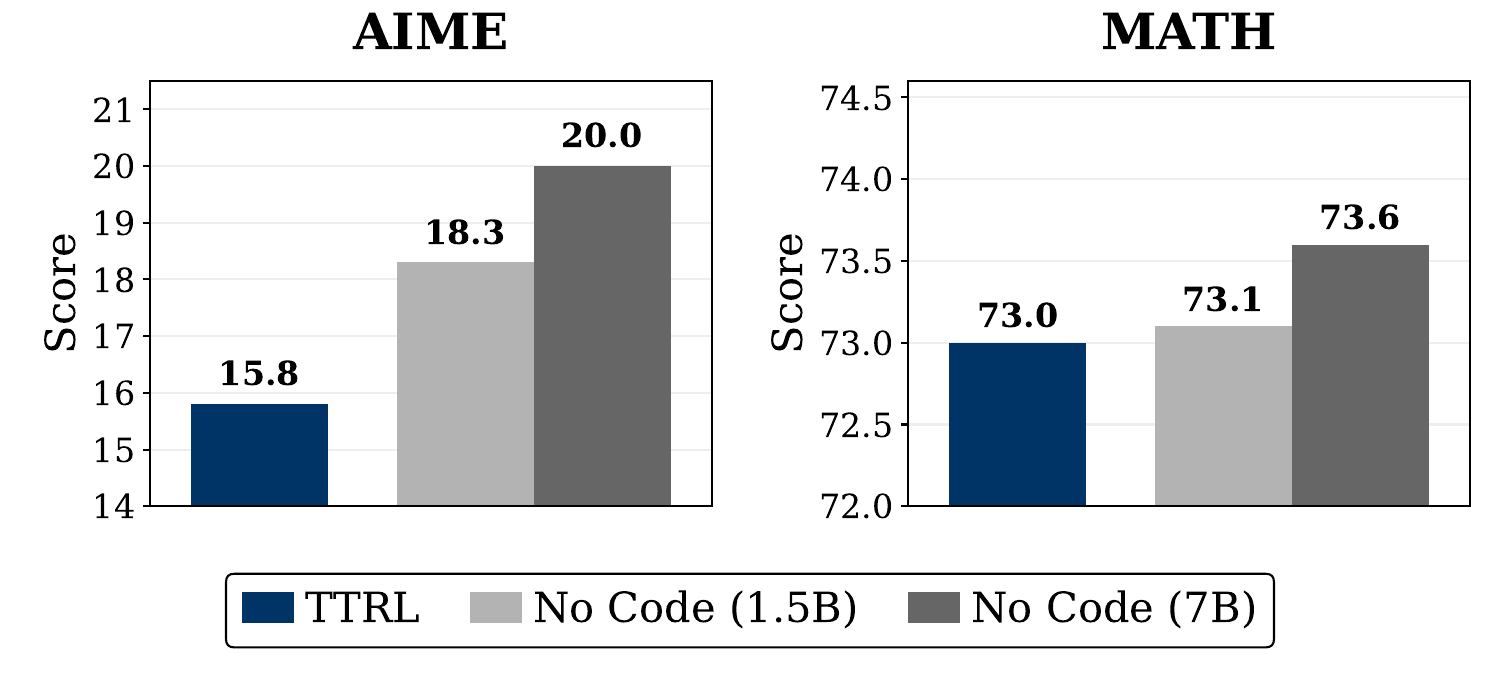}
    \caption{Ablating verification.}
    \label{fig:abl_llm_verifier}
  \end{subfigure}
  \hfill
  \begin{subfigure}[t]{0.495\columnwidth}
    \centering
    \includegraphics[width=\linewidth]{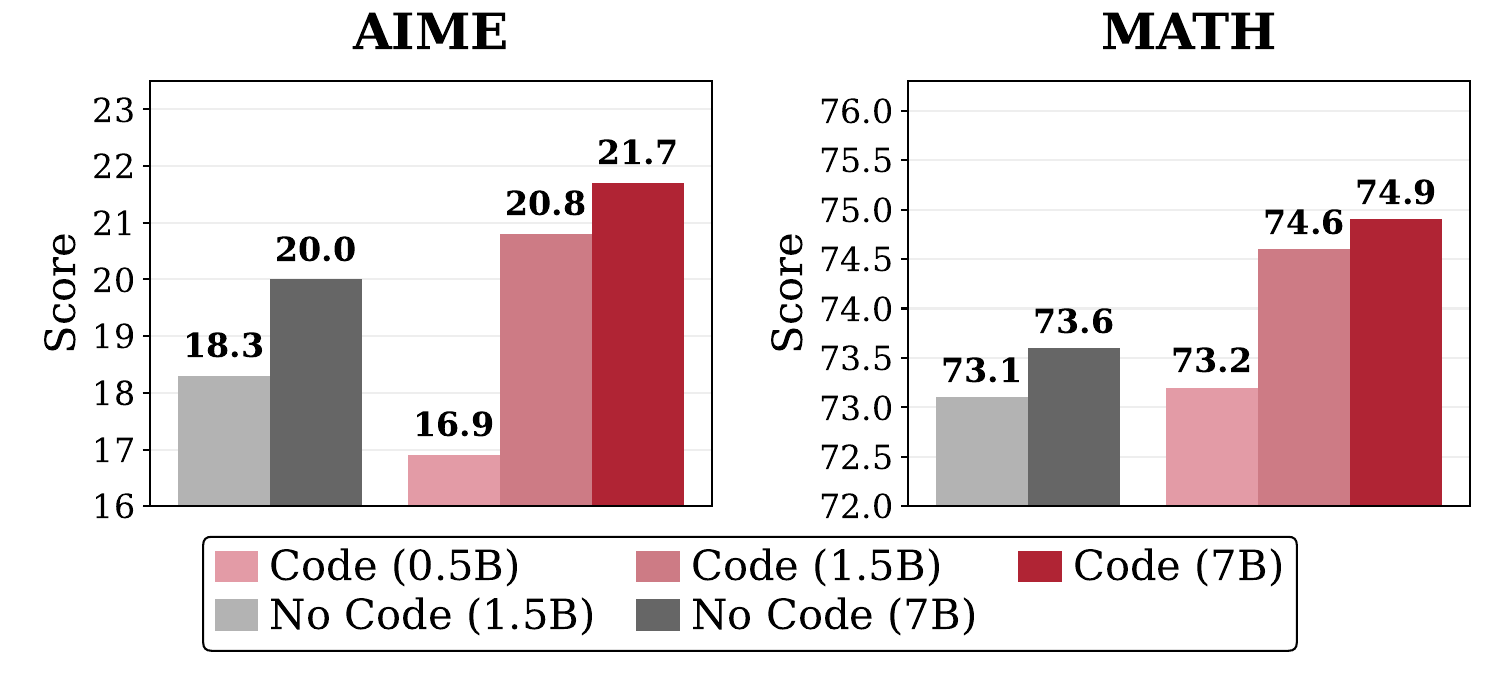}
    \caption{Ablating tool execution and verifier size.}
    \label{fig:abl_tool}
  \end{subfigure}
  \vspace{-4pt}
  \caption{\textbf{Ablation on \emph{verifier} and \emph{verification tool}.} Left: Adding an LLM verifier improves TTRL even without tool execution. Right: Code execution significantly strengthens verification.}
  \label{fig:ablation_two}
  \vspace{-8pt}
\end{figure}

\textbf{The Contribution of Test-Time Verification.} We isolate the effect of \emph{test-time verification} by comparing (i) vanilla TTRL against (ii) \method{} \emph{without} tool access, i.e., using LLM alone for verification. In \Cref{fig:abl_llm_verifier}, improvements are seen on both AIME and MATH with 1.5B or 7B verifier on 1.5B policy model. These results showcase that introducing verification on rollouts yields better online updates with improved reward estimation, even before introducing tool-assisted verification.

\textbf{The Contribution of Tool-Assisted Verification.} We next control for the \emph{tool execution} effect by comparing verification \method{} \emph{with} and \emph{without} tool execution. The results in \Cref{fig:abl_tool} show that the verification tool contributes a clear additional lift beyond verifier-only checking, improving AIME from $18.3$ to $20.8$ for the 1.5B verifier and from $20.0$ to $21.7$ for the 7B, confirming that executable evidence reduces verifier uncertainty and yields a more reliable verification signal.

\vspace{-8pt}
\noindent
\begin{minipage}[t]{0.72\textwidth}\vspace{0pt}
\textbf{Self-Verifier vs. External Verifier.} We further ablate the distillation effect by comparing \method{} with the policy model as self-verifier on AIME. Note that, the external verifier \texttt{Qwen-2.5-Coder-1.5B} is an even weaker model (1.7 on AIME). \Cref{tab:self_verifier} shows that self-verification beats \textsc{TTRL} on all three backbones. So the gain stems from \emph{verification itself} instead of distilling math knowledge from other model. Verification leads to flip rates on $5.40\%$ of AIME 2024, $10.75\%$ of AMC, and $12.87\%$ of MATH-500. 
\end{minipage}%
\hfill
\begin{minipage}[t]{0.25\textwidth}\vspace{0pt}
\centering
\captionsetup{type=table,font=scriptsize,skip=2pt}
\caption{Self vs.\ Ext.\ verifier.}
\label{tab:self_verifier}
\scriptsize
\setlength{\tabcolsep}{2pt}
\renewcommand{\arraystretch}{0.92}
\begin{tabular}{l|ccc}
\toprule
                                          & \makecell{Math\\1.5B} & \makecell{Llama\\3B} & \makecell{Qwen3\\4B} \\
\midrule
\textsc{TTRL}                             & 15.8 & 13.3 & 36.4 \\
\rowcolor{lightblue!25}Self               & 16.7 & 16.9 & 39.8 \\
\rowcolor{lightgreen!25}Ext.              & \textbf{20.8} & \textbf{17.1} & \textbf{40.0} \\
\bottomrule
\end{tabular}
\end{minipage}

\paragraph{The Contribution of Verification-Weighted Voting.}
\begin{wraptable}{r}{0.45\columnwidth}
  \centering
  \small
  \captionof{table}{\textbf{Verifier quality on policy rollouts.}}
  \label{tab:verifier_quality}
  \resizebox{\linewidth}{!}{%
  \begin{tabular}{ll|ccc|c}
    \toprule
    \textbf{Policy} & \textbf{Verifier} & \textbf{Acc.} & \textbf{Prec.} & \textbf{Rec.} & \textbf{$\Delta$ AIME} \\
    \midrule
    \texttt{Math-1.5B}    & \texttt{Coder-1.5B} & 73.2 & 25.0 & 51.5 & $+5.0$ \\
    \texttt{Llama-3.2-3B} & \texttt{Coder-1.5B} & 73.3 & 39.4 & 59.1 & $+3.8$ \\
    \texttt{Qwen-3-4B}    & \texttt{Coder-1.5B} & 75.0 & 38.2 & 40.7 & $+3.6$ \\
    \bottomrule
  \end{tabular}}
  \vspace{-2mm}
  \includegraphics[width=\linewidth]{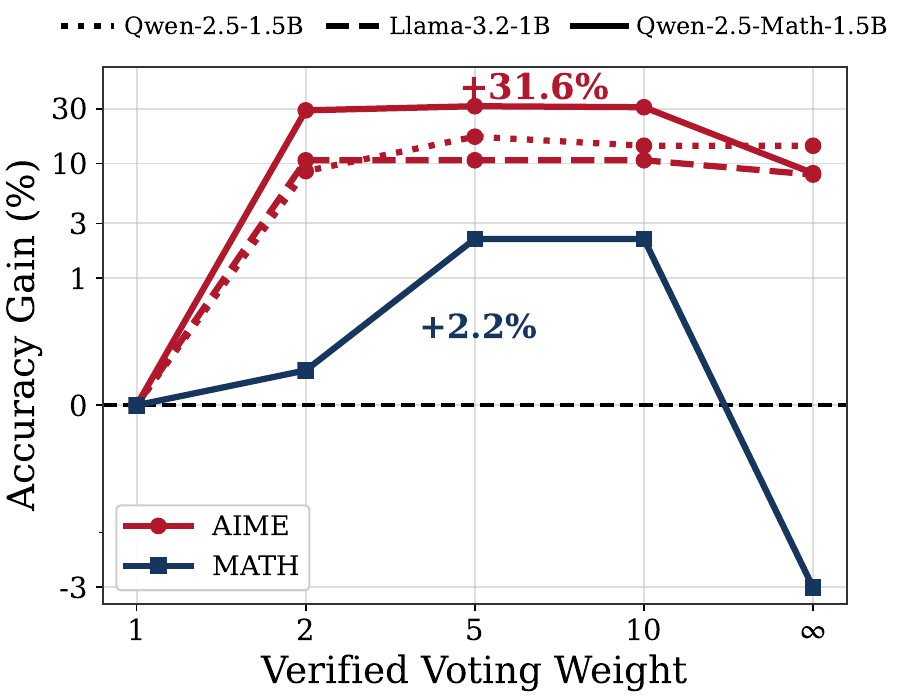}
  \captionof{figure}{The effect of the vote weight $\omega$.}
  \label{fig:abl_conf}
  \vspace{-8pt}
\end{wraptable}
We ablate the \emph{verification weight} $\omega$ in \method{} by sweeping its value, which controls the voting power of rollouts that pass the tool-verifier check. A weight of $\omega{=}1$ degenerates to majority-vote TTRL, and $\omega{\to}\infty$ approximates \emph{binary hard filtering} of unverified rollouts. As shown in \Cref{fig:abl_conf}, finite weights $\omega>1$ improve AIME accuracy across all three backbones. The largest gain occurs for \texttt{Qwen-2.5-Math-1.5B} at $\omega{=}5$ (AIME $+31.6\%$, MATH-500 $+2.2\%$), with similar performance at $\omega{=}10$; hard filtering is less reliable. This behavior arises because the verifier is itself noisy due to the stochastic nature of LLMs in rollout-to-code translation, although tool execution provides a deterministic check: \Cref{tab:verifier_quality} shows that the verifier accuracy with \texttt{Coder-1.5B} reaches only $73.2\%$, so hard filtering can propagate false positives and discard true positives, while $\omega{=}1$ ignores the verifier entirely. Therefore, tool-assisted verification should act as a \textbf{\emph{soft}} preference signal.

\changed{\textit{Sensitivity of the verification weight: } Every $\omega{>}1$ improves over \textsc{TTRL} generally, and moderate values plateau ($\omega{=}2$ is within $0.4$ points of $\omega{=}5$); only hard filtering degrades.}

\section{Discussions and Analysis}

\subsection{Q1: Why Does \method Work?}\label{sec:why_work}

\begin{takeaway}
\label{sec:data_syn}
\textcolor{darkgreen}{\itshape \method{} functions as a verified self-distillation loop that generates synthetic training data online.}
\end{takeaway}

Tool-use of reasoning models is learned from execution feedback and improved through interaction with the environment \citep{silver2025welcome}. In the test-time training loop, there's an open design choice: \textit{tools as policy actions} or \textit{tools as verification evidence.} The impressive contribution of tool verification in \method{} raises the question of whether tool access alone suffices, or whether the position in the loop also matters.

\begin{figure}[H]
  \centering
  \vspace{-2mm}
  \begin{subfigure}[b]{0.32\columnwidth}
    \centering
    \includegraphics[width=\linewidth]{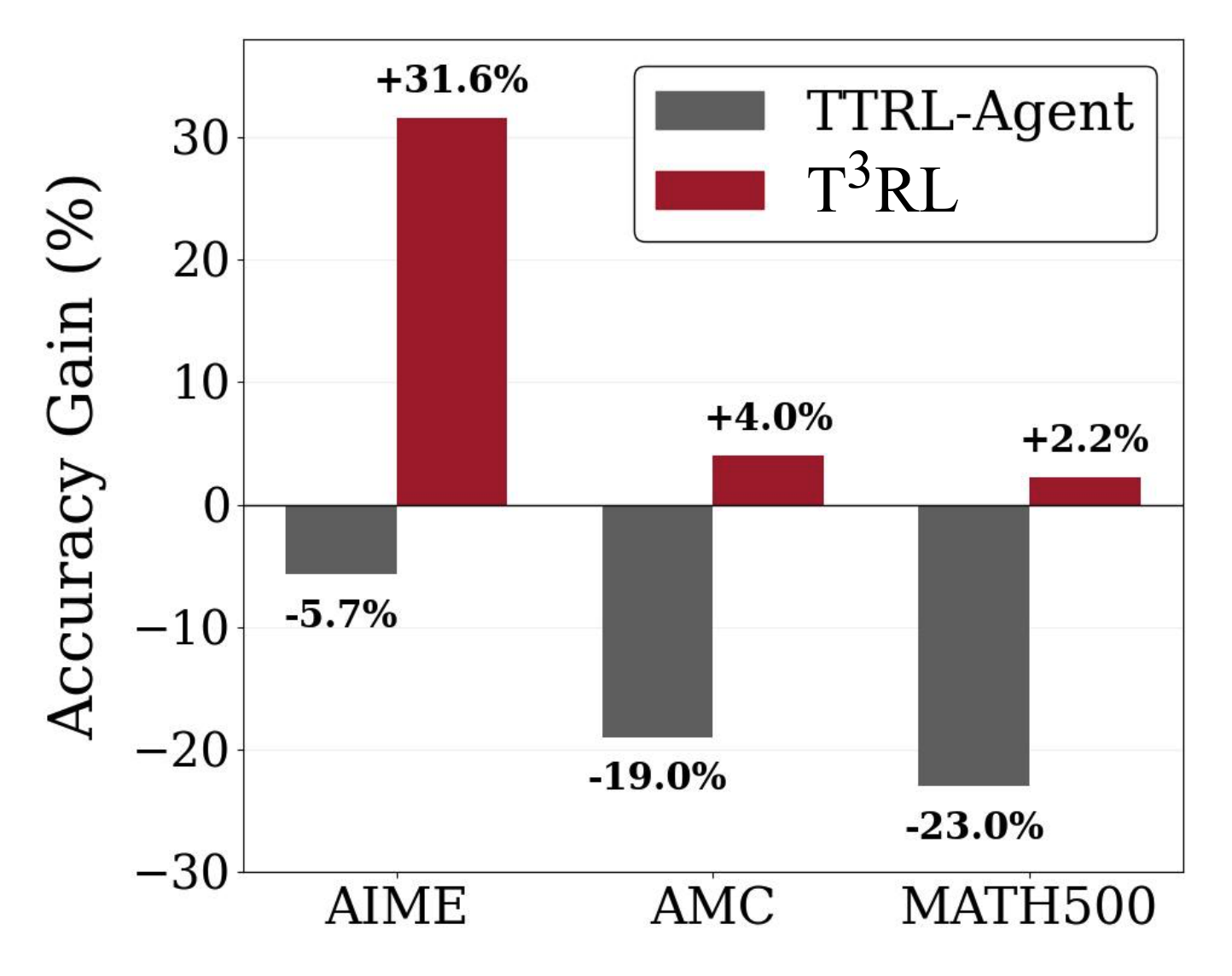}
    \vspace{-4mm}
    \caption{Tool in policy vs. in verification.}
    \label{fig:abl_tool_calling_vs_verification}
  \end{subfigure}
  \hfill
  \begin{subfigure}[b]{0.32\columnwidth}
    \centering
    \includegraphics[width=\linewidth]{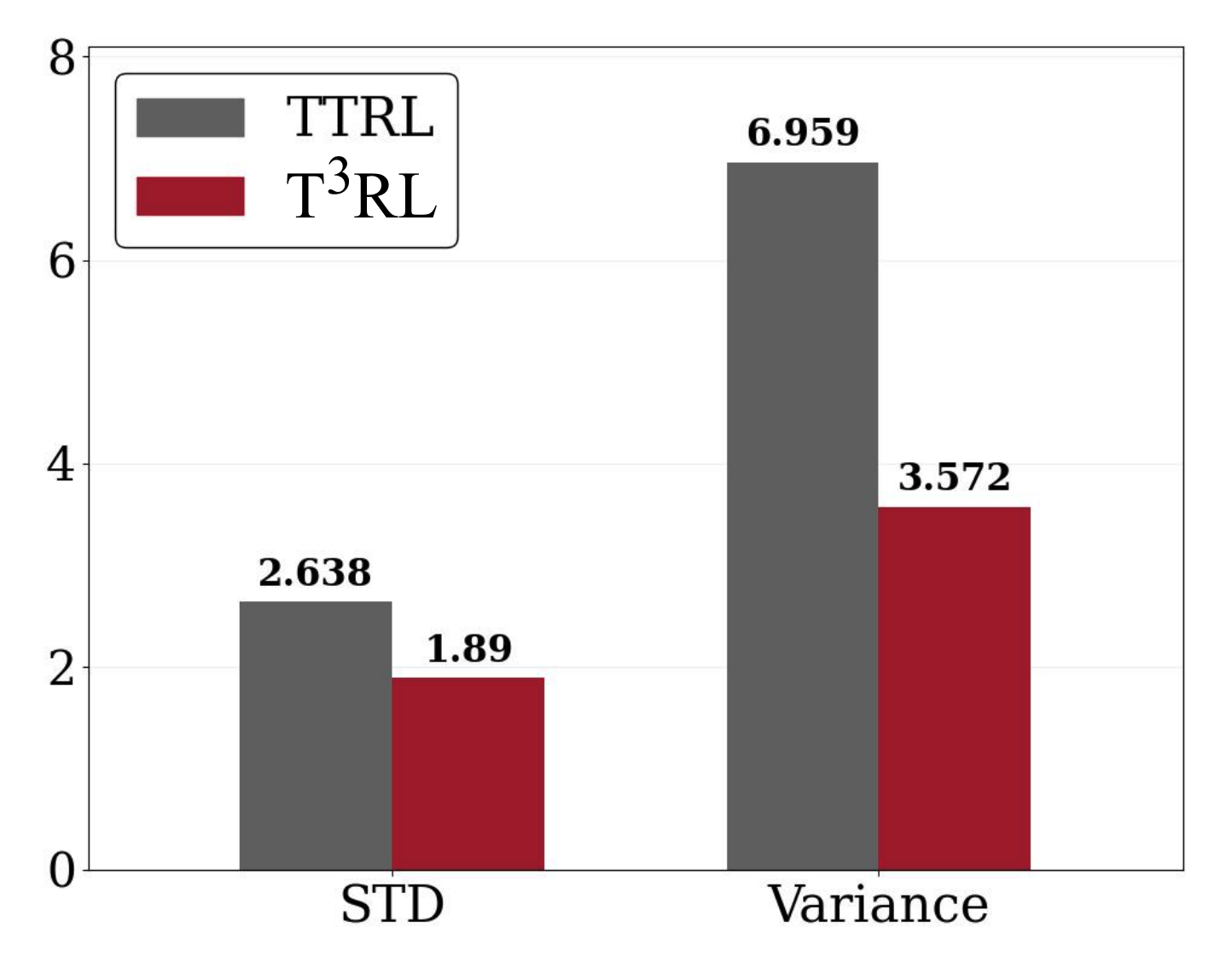}
    \vspace{-4mm}
    \caption{Training robustness.}
    \label{fig:robustness}
  \end{subfigure}
  \hfill
  \begin{subfigure}[b]{0.32\columnwidth}
    \centering
    \includegraphics[width=\linewidth]{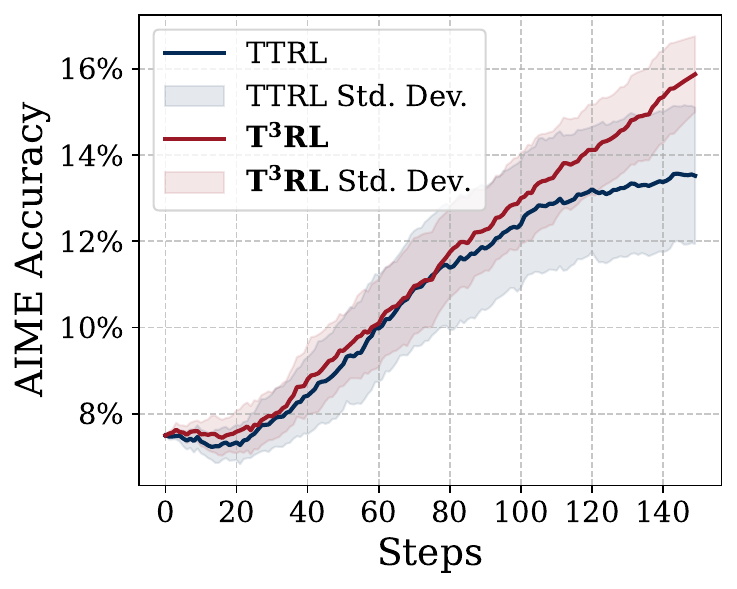}
    \vspace{-5.5mm}
    \caption{Training dynamics comparison.}
    \label{fig:dynamics}
  \end{subfigure}
  \caption{\textbf{Ablations on tool position and training robustness.}
  Left: allowing the \emph{policy} to call tools during rollouts (\textsc{TTRL-Agent}) can hurt performance, while \method{} improves by restricting tool use to the \emph{verifier} for reward shaping.
  Right: verifier-shaped rewards reduce run-to-run variability, indicating more stable optimization under unlabeled test-time RL.}
  \label{fig:abl_toolpos_and_robustness}
  \vspace{-6mm}
\end{figure}

\textit{Setting.} To further examine why \method{} works, we compare (i) \textsc{TTRL}; (ii) \textsc{TTRL-Agent}, which extends TTRL and grants the trained policy access to tool calling directly with majority voting over execution results, and (iii) \method{}, which trains a non-agentic policy with a tool-assisted \emph{verifier}.

\textit{Observation \& analysis.} \Cref{fig:abl_tool_calling_vs_verification} shows \textsc{TTRL-Agent} degrades, while \method{} yields growing positive gains with benchmark difficulty. We attribute \textsc{TTRL-Agent}'s failure to \emph{error-signal mixture}: tool calls inside rollouts conflate reasoning errors with tool-usage errors (e.g., malformed code generation or brittle execution artifacts), and self-consensus rewards amplify this noise.
In contrast, \method{} \emph{decouples} reasoning from tool execution: the verifier provides evidence \emph{after} generation and converts it into a verification-informed reward, turning rollouts into verified labeled training data on the fly: an implicit \textcolor{darkred}{\textit{verified online data synthesiser}} that stabilizes test-time learning.

\textbf{\method is more robust under tool-assisted verification voting.}
Tool verification further stabilizes training, as shown in \Cref{fig:dynamics}. Across multiple runs, \textsc{TTRL} exhibits noticeable run-to-run variability, consistent with the sensitivity of self-induced rewards to sampling noise and pseudo-label estimation. In contrast, \method{} anchors reward construction with tool verification, yielding substantially lower dispersion in peak performance in \Cref{fig:robustness}: both the standard deviation and variance of the best accuracy after $100$ steps. Overall, \method{} produces more consistent learning dynamics and more reliable accuracy on AIME.

\begin{table}[!ht]
\centering
\vspace{-2mm}
\begin{minipage}[t]{0.49\linewidth}\vspace{0pt}
  \centering
  \captionsetup{justification=raggedright,singlelinecheck=false}
  \caption{\textbf{Training computation.} AIME runs; arrows are relative to
  \textsc{TTRL}@$64$. \method achieves higher accuracy with less than half of the training cost, time and computations.}
  \label{tab:compute}
  \vspace{5pt}
  \scriptsize
  \setlength{\tabcolsep}{3pt}
  \renewcommand{\arraystretch}{1.05}
  {%
  \resizebox{\linewidth}{!}{%
  \begin{tabular}{l|cccc}
  \toprule
  \textbf{Method (rollout)} & \textbf{AIME}\,$\uparrow$ & \textbf{Tokens}\,$\downarrow$ & \textbf{Time\,(s)}\,$\downarrow$ & \textbf{TFLOPs}\,$\downarrow$ \\
  \midrule
  \textsc{TTRL}@$64$                          & 15.8 & 365K & 324 & 7.9K \\
  \rowcolor{lightgreen!25}\method{}@$16$     & \textbf{20.0}\,({\color{red}$\uparrow$27\%}) & \textbf{147K}\,({\color{blue}$\downarrow$60\%}) & \textbf{126}\,({\color{blue}$\downarrow$61\%}) & \textbf{3.0K}\,({\color{blue}$\downarrow$63\%}) \\
  \rowcolor{lightgreen!25}\method{}@$32$     & 20.8\,({\color{red}$\uparrow$32\%}) & 293K\,({\color{blue}$\downarrow$20\%}) & 218\,({\color{blue}$\downarrow$33\%}) & 5.9K\,({\color{blue}$\downarrow$25\%}) \\
  \method{}@$64$                              & 20.8\,({\color{red}$\uparrow$32\%}) & 586K\,({\color{red}$\uparrow$61\%}) & 406\,({\color{red}$\uparrow$26\%}) & 11.8K\,({\color{red}$\uparrow$50\%}) \\
  \bottomrule
  \end{tabular}}}
\end{minipage}
\hfill
\begin{minipage}[t]{0.49\linewidth}\vspace{0pt}
  \centering
  \captionsetup{justification=raggedright,singlelinecheck=false}
  \caption{\textbf{Inference computation.} Per-query costs; arrows are relative
  to \method{}.}
  \label{tab:nonparametric}
  \scriptsize
  \setlength{\tabcolsep}{3pt}
  \renewcommand{\arraystretch}{1.05}
  {%
  \resizebox{\linewidth}{!}{%
  \begin{tabular}{l|c|ccc}
  \toprule
  \textbf{Method} & \textbf{Inf.\ Tokens}\,$\downarrow$ & \textbf{AIME} & \textbf{AMC} & \textbf{MATH} \\
  \midrule
  Base                              & 1{,}630 & 7.7\,({\color{red}$\downarrow$63\%}) & 28.6\,({\color{red}$\downarrow$44\%}) & 32.7\,({\color{red}$\downarrow$56\%}) \\
  \textsc{TTRL}                     & 1{,}630 & 15.8\,({\color{red}$\downarrow$24\%}) & 48.9\,({\color{red}$\downarrow$4\%}) & 73.0\,({\color{red}$\downarrow$2\%}) \\
  \rowcolor{lightgreen!25}\textbf{\method{} (ours)} & \textbf{1{,}630} & \textbf{20.8} & \textbf{50.9} & \textbf{74.6} \\
  \midrule
  {\color{diffblue}Test-time majority ($N{=}16$)} & {\color{diffblue}61K\,({\color{red}$\uparrow$37$\times$})} & {\color{diffblue}19.2\,({\color{red}$\downarrow$8\%})} & {\color{diffblue}49.4\,({\color{red}$\downarrow$3\%})} & {\color{diffblue}65.0\,({\color{red}$\downarrow$13\%})} \\
  {\color{diffblue}Tool-verified majority ($N{=}16$)} & {\color{diffblue}96K\,({\color{red}$\uparrow$59$\times$})} & {\color{diffblue}20.8\,({\color{gray}$\rightarrow$0\%})} & {\color{diffblue}51.5\,({\color{blue}$\uparrow$1\%})} & {\color{diffblue}65.5\,({\color{red}$\downarrow$12\%})} \\
  {\color{diffblue}T1 ($N{=}16,V{=}4$)} & {\color{diffblue}200K\,({\color{red}$\uparrow$123$\times$})} & {\color{diffblue}20.0\,({\color{red}$\downarrow$4\%})} & {\color{diffblue}52.1\,({\color{blue}$\uparrow$2\%})} & {\color{diffblue}66.3\,({\color{red}$\downarrow$11\%})} \\
  {\color{diffblue}T1 + PRM ($N{=}16,V{=}4$)} & {\color{diffblue}200K\,({\color{red}$\uparrow$123$\times$})} & {\color{diffblue}20.4\,({\color{red}$\downarrow$2\%})} & {\color{diffblue}52.7\,({\color{blue}$\uparrow$4\%})} & {\color{diffblue}67.9\,({\color{red}$\downarrow$9\%})} \\
  Buffer-of-Thoughts                & 2{,}120\,({\color{red}$\uparrow$30\%}) & 9.6\,({\color{red}$\downarrow$54\%}) & 34.9\,({\color{red}$\downarrow$31\%}) & 38.6\,({\color{red}$\downarrow$48\%}) \\
  BoN-RR ($N{=}64$)                 & 243K\,({\color{red}$\uparrow$148$\!\times$}) & 21.0\,({\color{blue}$\uparrow$1\%}) & 51.8\,({\color{blue}$\uparrow$2\%}) & 51.7\,({\color{red}$\downarrow$31\%}) \\
  
  \bottomrule
  \end{tabular}}}
\end{minipage}
\vspace{-4pt}
\end{table}

\textbf{Allocating test-time computation wisely: verification beats brute-force
scaling.}\label{sec:compute}
How should the test-time budget be split between (i)~\emph{sampling more
rollouts} for self-consensus and (ii)~\emph{verifying rollouts} to make each
sample more informative? We find two things.
\emph{First, \method{} is cheaper during training ($60\%$ less) to achieve better accuracy than \textsc{TTRL}.}
Holding all else fixed and accounting for both
policy and verifier costs, \method{}@$r{=}16$ already surpasses
\textsc{TTRL}@$64$ using $59.8\%$ fewer tokens, $61.3\%$ less wall-clock
time, and $62.5\%$ fewer TFLOPs (\Cref{tab:compute}); sandbox execution adds
only ${\sim}49$\,ms per snippet ($<\!1\%$ of training).
\emph{Second, \method{} is cheaper during inference ($123$--$200\times$ less) than verifier-based TTS to achieve matched accuracy.}
Comparing with non-parametric alternatives
(\Cref{tab:nonparametric}): Buffer-of-Thoughts~\citep{yang2024buffer}, a
memory-based system that reuses successful rollouts, cannot match even
\textsc{TTRL}, while the verifier-guided test-time scaling approach, best-of-$N$
reranking (BoN-RR), approaches \method{}'s accuracy only by inflating per-query
inference tokens by $148\!\times$ and paying $\mathcal{O}(2N)$ at every test
instance. \changed{Other verifier-aware TTS baselines are no cheaper:
T1~\citep{ICLR2026_776a5f2c} reaches $66.3$ and
T1+PRM~\citep{wang-etal-2024-math} reaches $67.9$ on MATH-500, both at $200$K
tokens per query.} \method{} instead
amortises the verifier's signal into the policy weights once at training and
infers at $\mathcal{O}(1)$, converting test-time compute into efficient
parameter updates and avoiding re-paying the inference cost on every query.

\needspace{8\baselineskip}
\subsection{Q2: Can \method{} generalize?}
\label{sec:domain_generalization}

\textbf{Generalization beyond math: logical reasoning}
To test whether \method{} depends on mathematical reasoning or Python
execution, we evaluate it on held-out two-character Knights-and-Knaves
puzzles, where knights always tell the truth and knaves always lie. The verifier
represents knights as \texttt{true} and knaves as \texttt{false}, translates policy
solutions to the problem into Satisfiability Modulo Theories
 (SMT)~\citep{barrett2018satisfiability} constraints using SMT-LIB2, and checks them with
Z3 solver execution~\citep{xie2024memorization,demoura2008z3}.

\begin{table}[h]
\centering
\vspace{-5pt}
\captionsetup{skip=1pt}
\caption{\changed{Non-math generalization on Knights-and-Knaves puzzles with an SMT-LIB2/Z3 example.}}
\label{tab:knights_knaves}
\begin{minipage}[t][46pt][c]{0.53\linewidth}\vspace{0pt}
\centering
\scriptsize
{\color{diffblue}
\resizebox{0.87\linewidth}{!}{%
\begin{tabular}{lccc}
\toprule
\textbf{Method} & \textbf{Base model} & \textbf{Accuracy} & $\Delta$ \textbf{vs. base} \\
\midrule
Base & Qwen-2.5-VL-1.5B & $52.13\pm1.3$ & -- \\
TTRL & Qwen-2.5-VL-1.5B & $48.58\pm1.3$ & $-3.55$ \\
\method{} & Qwen-2.5-VL-1.5B & $\mathbf{78.50\pm1.1}$ & $\mathbf{+26.38}$ \\
\bottomrule
\end{tabular}}}
\end{minipage}
\hfill
\begin{minipage}[t][46pt][c]{0.45\linewidth}\vspace{0pt}
\centering
\begin{tikzpicture}[baseline=(current bounding box.center)]
  \node[
    fill=flowPaleGreen, rounded corners=2pt,
    minimum width=28pt, minimum height=10pt,
    align=center, inner sep=1pt, font=\tiny\bfseries, text=black
  ] (query) at (27pt,13pt) {Query};
  \node[
    fill=gray!15, rounded corners=2pt,
    minimum width=54pt, minimum height=31pt, text width=51pt,
    align=center, inner sep=1.5pt, font=\tiny, text=black
  ] (kk) at (27pt,-6pt) {A says: ``B is a\\[-1pt]
    \textcolor{barRed}{knave}.''\\[-1pt]
    B says: ``A and I\\[-1pt]
    are different.''};
  \node[
    fill=flowPaleGreen, rounded corners=2pt,
    minimum width=34pt, minimum height=10pt,
    align=center, inner sep=1pt, font=\tiny\bfseries, text=black
  ] (verifier) at (96pt,14pt) {Verifier};
  \node[
    fill=flowTeal, rounded corners=2pt,
    minimum width=76pt, minimum height=31pt, text width=73pt,
    align=center, inner sep=1.5pt, font=\tiny, text=white
  ] (smt) at (96pt,-6pt) {\textbf{SMT-LIB2 constraints}\\[-1pt]
    \texttt{(assert (= A (not B)))}\\[-1pt]
    \texttt{(assert (= B (xor A B)))}};
  \node[
    fill=flowPaleGreen, rounded corners=2pt,
    minimum width=24pt, minimum height=10pt,
    align=center, inner sep=1pt, font=\tiny\bfseries, text=black
  ] (z3label) at (157pt,13pt) {Z3 result};
  \node[
    fill=gray!15, rounded corners=2pt,
    minimum width=42pt, minimum height=31pt, text width=39pt,
    align=center, inner sep=1.5pt, font=\tiny, text=black
  ] (z3) at (157pt,-6pt) {\textcolor{darkgreen}{\ding{51}}\;\textbf{sat}\\[-1pt]
    \texttt{A=false} (knave)\\[-1pt]
    \texttt{B=true} (knight)};
  \draw[->, color=flowTeal, line width=0.6pt] (kk.east) -- (smt.west);
  \draw[->, color=flowTeal, line width=0.6pt] (smt.east) -- (z3.west);
\end{tikzpicture}
\end{minipage}
\vspace{-8pt}
\end{table}

\method{} improves accuracy with a gain of $26.38$ points over the base model
(\Cref{tab:knights_knaves}) while TTRL decreases. This result supports the narrower claim that
\method{} can generalize when an external tool provides a reliable correctness
check, for example, on logical reasoning intensive tasks such as SMT solving, theorem proving, and program verification.

\textbf{Generalization across-dataset. }
Each row adapts the policy at test time on one source benchmark and evaluates it on all
three targets, so off-diagonal entries measure transfer to unseen benchmarks.
\method{} improves all nine train--test pairs (mean $+3.0$ points;
\Cref{tab:cross_dataset}). Mean run-to-run SD decreases by $8.4\%$
($2.38\!\to\!2.18$).

\begin{table}[h]
\centering
\vspace{-8pt}
\captionsetup{skip=1pt}
\caption{\changed{Cross-dataset adaptation (A/C/M: AIME/AMC/MATH-500). Inset: accuracy gain.}}
\label{tab:cross_dataset}
\scriptsize
\setlength{\tabcolsep}{4pt}
\renewcommand{\arraystretch}{0.9}
\begin{tabular}{@{}c@{\hspace{16pt}}c@{}}
{\color{diffblue}%
\begin{tabular}{lccc}
\toprule
\textbf{Adaptation setting} & \textbf{AIME 2024} & \textbf{AMC} & \textbf{MATH-500} \\
\midrule
\rowcolor{gray!12}TTRL adapted on AIME & $15.8\pm2.1$ & $42.9\pm2.4$ & $65.5\pm2.4$ \\
\rowcolor{gray!12}\method{} adapted on AIME & $\mathbf{20.8\pm1.6}$ & $45.4\pm2.5$ & $68.0\pm2.0$ \\
\rowcolor{white}TTRL adapted on AMC & $11.5\pm1.6$ & $48.9\pm1.6$ & $66.6\pm1.4$ \\
\rowcolor{white}\method{} adapted on AMC & $\mathbf{15.1\pm2.0}$ & $\mathbf{50.9\pm2.7}$ & $\mathbf{70.4\pm1.7}$ \\
\rowcolor{gray!12}TTRL adapted on MATH-500 & $10.8\pm3.4$ & $43.2\pm2.7$ & $73.0\pm3.8$ \\
\rowcolor{gray!12}\method{} adapted on MATH-500 & $\mathbf{14.7\pm1.9}$ & $\mathbf{45.7\pm2.0}$ & $\mathbf{74.6\pm3.2}$ \\
\bottomrule
\end{tabular}}
&
\resizebox{68pt}{!}{%
\begin{tikzpicture}[baseline=(current bounding box.center)]
  \node[font=\tiny\bfseries] at (25pt,15pt) {Accuracy gain (pp)};
  \foreach \x/\lab in {14/A,32/C,50/M} {
    \node[font=\tiny] at (\x pt,8pt) {\lab};
  }
  \foreach \y/\lab in {0/A,-9/C,-18/M} {
    \node[font=\tiny,anchor=east] at (4pt,\y pt) {\lab};
  }
  \node[fill=flowTeal!80,text=white,minimum width=17pt,minimum height=9pt,font=\tiny\bfseries] at (14pt,0pt) {$+5.0$};
  \node[fill=flowTeal!55,minimum width=17pt,minimum height=9pt,font=\tiny\bfseries] at (32pt,0pt) {$+2.5$};
  \node[fill=flowTeal!55,minimum width=17pt,minimum height=9pt,font=\tiny\bfseries] at (50pt,0pt) {$+2.5$};
  \node[fill=flowTeal!68,text=white,minimum width=17pt,minimum height=8pt,font=\tiny\bfseries] at (14pt,-9pt) {$+3.6$};
  \node[fill=flowTeal!48,minimum width=17pt,minimum height=8pt,font=\tiny\bfseries] at (32pt,-9pt) {$+2.0$};
  \node[fill=flowTeal!70,text=white,minimum width=17pt,minimum height=8pt,font=\tiny\bfseries] at (50pt,-9pt) {$+3.8$};
  \node[fill=flowTeal!72,text=white,minimum width=17pt,minimum height=8pt,font=\tiny\bfseries] at (14pt,-18pt) {$+3.9$};
  \node[fill=flowTeal!55,minimum width=17pt,minimum height=8pt,font=\tiny\bfseries] at (32pt,-18pt) {$+2.5$};
  \node[fill=flowTeal!42,minimum width=17pt,minimum height=8pt,font=\tiny\bfseries] at (50pt,-18pt) {$+1.6$};

  \node[font=\tiny] at (27pt,-36pt) {\textcolor{flowTeal}{\rule{5pt}{5pt}} \method{} better};
\end{tikzpicture}}
\\
\end{tabular}
\vspace{-10pt}
\end{table}

\needspace{5\baselineskip}
\subsection{Q3: What boosts \method{}, and when might it fail?}
\label{sec:boost_fail}

\textbf{Boosting performance: stronger verifiers and larger rollout budgets help.}
Scaling the verifier from $0.5$B to $7$B on \texttt{Qwen-Math-2.5}
consistently improves \method{} across benchmarks (AIME $20.8\!\to\!21.7$,
AMC $50.9\!\to\!51.5$, MATH-500 $74.4\!\to\!74.9$;
\Cref{tab:verifier_quality}): stronger verifiers produce more reliable
rollout-to-code translation and verification-aware reward signals. Increasing
the rollout budget $N\!\in\!\{16,32,64\}$ also improves \method{}
monotonically (\Cref{tab:compute}), as a larger candidate set raises solution
diversity and concentrates vote mass on a more reliable pseudo-label.
\Cref{fig:mvflip} shows the resulting effect: \method{} flips the majority
label via tool verification when an incorrect calculation dominates.

\textbf{Limitations: noisy verifiers and easy tasks limit gains.}
\method{} requires $\mathcal{V}$ to provide a meaningful correctness signal.
When $\mathcal{V}$ is underpowered (e.g., \texttt{Qwen-0.5B} with bare-minimum
coding ability), noisy tool calls cause verification-aware voting to misweight
rollouts and add a second stochastic layer on top of self-consensus, so the
pseudo-label $\widetilde{y}^{*}$ becomes less stable than vanilla majority
voting. Experimental results and failure-case studies are provided in
Appendix~\ref{appx:failure_cases}. On easy tasks where rollouts are already
accurate and consistent, verification aligns with the consensus without
changing the pseudo-label distribution, so gains over \textsc{TTRL} are small.

\section{Conclusion}
\label{sec:conclusion}

We propose \method{}, introducing \emph{test-time verification} to the test-time reinforcement learning framework that learns from unlabeled test data by suppressing spurious rewards with tool verification. Experiments across heterogeneous backbones and math benchmarks show consistent gains of \emph{tool verification}. Overall, \method{} positions test-time RL as \emph{verified online data synthesis}: sampled rollouts become reliable training instances once verified with executable evidence, enabling more stable verifiable test-time training. Future work will broaden to
domain-specific tool verification.

\section*{Impact Statement}
This paper aims to advance the field of machine learning.
Our work introduces test-time verification as a mechanism
for stabilizing self-evolution in large reasoning models by reducing error reinforcement and mitigating self-consistencydriven failure modes. If deployed responsibly, such verification can improve reliability and robustness in high-stakes applications by encouraging models to seek and check external
evidence rather than relying solely on internal consistency.
Potential risks include over-reliance on imperfect verifiers
and the possibility that verification pipelines inherit biases or
vulnerabilities from their underlying tools and data sources.
To mitigate these risks, future work should investigate improvements to verifiers and robustness to adversarial or
noisy feedback. More broadly, the proposed framework is
modular and can incorporate improved test-time verification
methods as they become available.

\bibliographystyle{plainnat}
\bibliography{t3rl_neurips2026}

\appendix

\section{Implementation details: Pseudo-Code}
\label{app:code}
\noindent\begin{minipage}{\linewidth}
\begin{lstlisting}[
  rulecolor=\color{black},
  basicstyle=\color{diffblue}\ttfamily\small,
  label={lst:t3rl_pseudocode},
  caption={Pseudo-code for \method{}},
  abovecaptionskip=2pt,
  belowcaptionskip=7pt,
  language=Python,
]
from collections import defaultdict

def t3rl_reward_fn(x, policy, verifier, sandbox, N, omega):
    Y = policy.sample_rollouts(x, n=N) 
    vote, A = defaultdict(float), []

    for y in Y:
        a_hat = extract_answer(y)
        # (1) verifier generates code
        code = verifier.generate(x, y)
        # (2) sandbox executes the code and returns evidence
        t = sandbox.execute(code)
        # (3) verifier returns only a validity flag
        v = verifier.judge(x, y, a_hat, t)       # v in {0,1}
        vote[a_hat] += (1.0 if v == 0 else omega)
        A.append(a_hat)

    y_star  = max(vote, key=vote.get)      # weighted majority pseudo-label
    rewards = [1.0 if a_hat == y_star else 0.0 for a_hat in A]
    return y_star, rewards
\end{lstlisting}
\end{minipage}

\FloatBarrier
\section{Verifier System Prompt}
\label{appx:verifier_prompt}

\begin{figure}[h]
    \centering
    \includegraphics[width=0.7\linewidth]{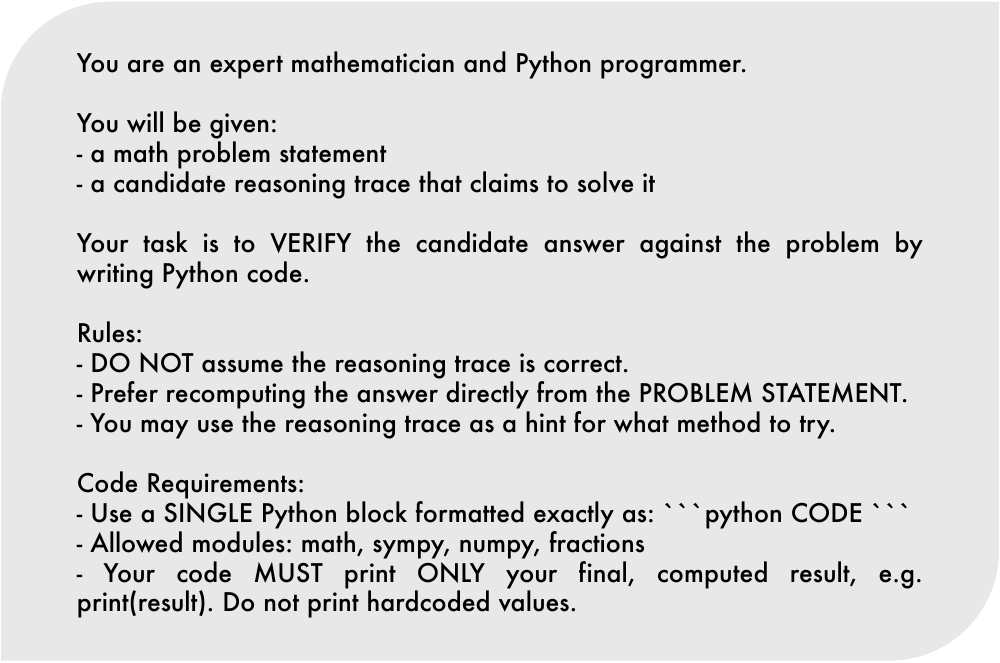}
    \caption{System Prompt for the Verifier.}
    \label{fig:verifier_prompt}
\end{figure}

The system prompt used to guide the external LLM verifier is shown in \Cref{fig:verifier_prompt}. We carefully designed the instructions to ensure reliable, independent, and easily parsable tool-assisted verification. The key design choices in our prompt include:

\begin{itemize}
    \item \textbf{Role Assignment:} Instructing the model to act as an ``expert mathematician and Python programmer'' sets a strong prior for generating rigorous, high-quality script formulations.
    \item \textbf{Independent Recomputation:} By explicitly stating ``DO NOT assume the reasoning trace is correct'' and ``Prefer recomputing the answer directly,'' we mitigate confirmation bias. This prevents the verifier from blindly translating a flawed reasoning trace into code, forcing it to independently verify the underlying logic based on the original problem statement.
    \item \textbf{Trace as a Hint:} Allowing the verifier to use the candidate trace as a ``hint'' helps it navigate complex problems. It can leverage the policy model's mathematical intuition without being strictly bound to its (potentially flawed) arithmetic execution.
\end{itemize}

\FloatBarrier
\section{Failure Case: Small Verifiers}
\label{appx:failure_cases}

\begin{figure}[t]
    \centering
    \begin{subfigure}[t]{0.45\textwidth}
        \centering
        \includegraphics[width=\textwidth]{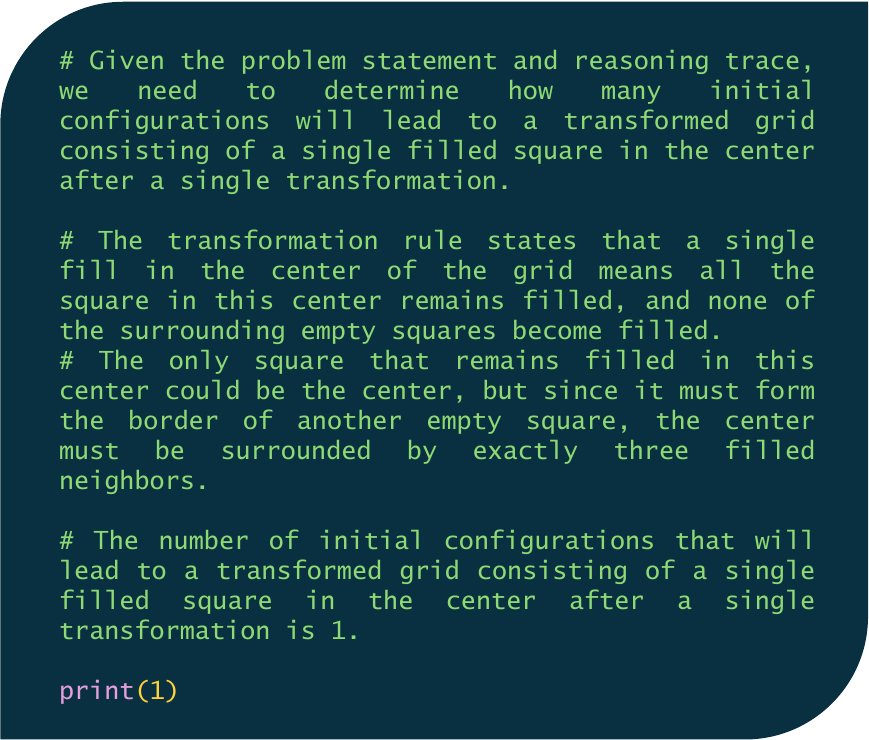}
        \caption{Blind copying and hardcoded outputs. The model hallucinates reasoning in code comments and simply prints the unverified final answer given in the reasoning trace.}
        \label{fig:fail_comments}
    \end{subfigure}
    \hspace{3mm}
    \begin{subfigure}[t]{0.45\textwidth}
        \centering
        \includegraphics[width=\textwidth]{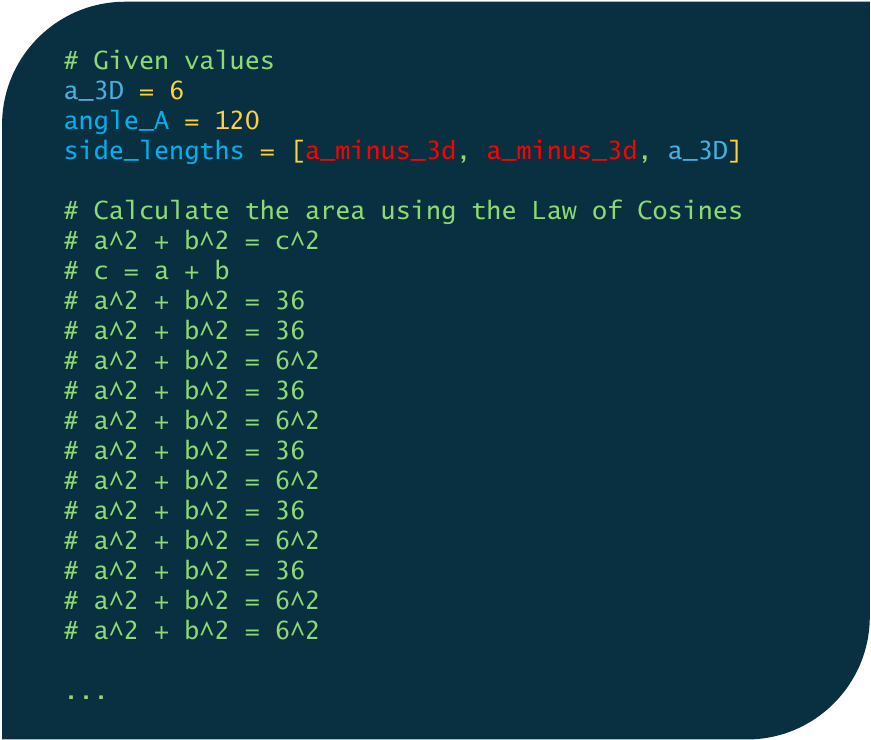}
        \caption{Formatting and compilation errors. The model fails to generate executable Python syntax and produces endless comments, resulting in compilation errors.}
        \label{fig:fail_compile}
    \end{subfigure}

    \vspace{2mm}
    \caption{Qualitative examples of the two primary failure modes encountered when deploying an undersized verifier.}
    \label{fig:verifier_failure_modes}
\end{figure}

As quantitative results in \Cref{tab:failure_cases_05b} demonstrate, deploying an undersized verifier like \texttt{Qwen-Coder-0.5B} actively degrades performance compared to the standard TTRL baseline. This performance drop stems from the small model's limited capacity to strictly adhere to the verification instructions, which injects noise rather than reliable evidence into the reward signal. Qualitative analysis of the generated verification scripts reveals two primary failure modes:

\begin{itemize}
    \item \textbf{Blind Copying and Hardcoded Outputs:} Despite explicit instructions in the system prompt to "DO NOT assume the reasoning trace is correct" and to strictly outputting hardcoded values, the 0.5B verifier frequently exhibits severe instruction-following failures. Instead of formulating an independent computational check, the model often defers entirely to the provided candidate trace. It typically hallucinates pseudo-reasoning within Python comments and bypasses actual computation, yielding a script that merely executes a hardcoded print statement of the trace's final answer (see \Cref{fig:fail_comments}). This confirmation bias neutralizes the benefits of executable verification, creating a false-positive signal that simply reinforces the unverified consensus.
    \item \textbf{Formatting and Compilation Errors:} A secondary, yet pervasive, issue is the small model's inability to consistently follow structural guidelines. The 0.5B verifier struggles to maintain valid Python syntax and adhere to the strict code-block formatting constraints required for automated extraction. This manifests as an increased frequency of compilation and execution errors, driven by missing import statements, malformed block delimiters, syntactical hallucinations, or endless comments (see \Cref{fig:fail_compile}). Consequently, the verification tool returns execution failures rather than reliable validity checks, further destabilizing the reward estimation process.
\end{itemize}

Combined, these limitations demonstrate that \method requires a minimum threshold of verifier capacity to function effectively; below this threshold, the verifier acts as an additional source of stochastic noise rather than a reliable grounding mechanism.

\begin{table}[h]
    \centering
    \resizebox{0.7\linewidth}{!}{
    \begin{tabular}{l|cccc}
    \toprule
    \textbf{Model / Method} & \textbf{AIME 2024} & \textbf{AMC} & \textbf{MATH-500} & \textbf{Avg} \\
    \midrule
    Qwen-2.5-0.5B Vanilla (Baseline) & 0.0 & 4.8 & 7.9 & 4.2 \\
    \quad w/ TTRL & 0.4 & 12.0 & 34.6 & 15.7 \\
    \rowcolor{lightgreen!25}\quad \textbf{w/ \method{} (0.5B verifier)} & \textbf{0.0} & \textbf{10.8} & \textbf{32.0} & \textbf{14.3} \\
    \rowcolor{lightblue!100}\quad $\Delta$ (\method{} $-$ TTRL) & {\color{blue}-0.4} & {\color{blue}-1.2} & {\color{blue}-2.6} & {\color{blue}-1.4} \\
    \rowcolor{lightgreen!25}\quad Rel.\ (\% over TTRL) & {\color{blue}$\downarrow$100.0\%} & {\color{blue}$\downarrow$10.0\%} & {\color{blue}$\downarrow$7.5\%} & {\color{blue}$\downarrow$8.9\%} \\
    \bottomrule
    \end{tabular}
    }
    \vspace{2mm}
    \caption{Performance comparison illustrating failure cases with a weak (0.5B) verifier.}
    \label{tab:failure_cases_05b}
\end{table}


\end{document}